\documentclass{mlgarticle}

\usepackage{amsmath,amssymb,amsfonts,amsbsy,amsthm}
\usepackage{latexsym}
\usepackage{graphicx,psfrag,epsfig,subfigure}
\usepackage{multicol,multirow}

\usepackage[numbers,sort]{natbib}
\usepackage{longtable}
\usepackage{pdflscape}

\usepackage[ruled,commentsnumbered]{algorithm2e}

\input{def.set}

\MLGreportno{001}
\MLGreportyear{2014}
\title{\bf Convex Optimization for Binary Classifier Aggregation in Multiclass Problems
\footnote{Appeared in Proceedings of the 2014 SIAM International Conference on Data Mining (SDM 2014).} }
\subtitle{}
\author{Sunho Park$^{\S}$,  TaeHyun Hwang$^{\S}$,  Seungjin Choi$^\dag$\\}
\institute{$^{\S}$ Clinical Sciences, University of Texas Southwestern Medical Center , Dallas, TX 75390, USA.\\
Email: \{sunho.park, taehyun.hwang\}@utsouthwestern.edu  \\ \\
$^\dag$ \MLGaddress\\
Email: seungjin@postech.ac.kr}
\shorttitle{Convex Binary Classifier Aggregation}
\shortauthor{S. Park et al.}

\abstract{
Multiclass problems are often decomposed into multiple binary problems
that are solved by individual binary classifiers whose results are integrated into a final answer.
Various methods, including all-pairs (APs), one-versus-all (OVA), and error correcting output code (ECOC),
have been studied,  to decompose multiclass problems into binary problems.
However, little study has been made to optimally aggregate binary problems to determine a final answer to
the multiclass problem.
In this paper we present a convex optimization method for an optimal aggregation of binary classifiers
to estimate class membership probabilities in multiclass problems.
We model the class membership probability as a softmax function which takes
a conic combination of discrepancies induced by individual binary classifiers, as an input.
With this model, we formulate the regularized maximum likelihood estimation
as a convex optimization problem, which is solved by the primal-dual interior point method.
Connections of our method to large margin classifiers are presented, showing that
the large margin formulation can be considered as a limiting case of our convex formulation.
Numerical experiments on synthetic and real-world data sets demonstrate that
our method outperforms existing aggregation methods as well as direct methods,
in terms of the classification accuracy and the quality of class membership probability estimates.
}

\acknowledgement{
}

\begin{document}

\section{Introduction}
\label{sec:introduction}

Multiclass classification is an important supervised learning problem,
the goal of which is to assign data points to a finite set of $K$ classes, which is solved by one of
two different approaches (direct and indirect methods).
Direct approach involves constructing a discriminant function directly for the multiclass problem.
For example, the multiclass SVM \cite{WestonJ99esann,CrammerK2001jmlr} models a $K$-way classifier
which directly separates the correct class label from the rest of class labels in the large margin framework.
Alternatively, in indirect approach, one decomposes the multiclass problem into multiple binary classification problems
which are solved by individual binary classifiers whose results are integrated into a final answer.
All-pairs (APs) and one-versus-all (OVA) are well-known methods for decomposing multiclass problems into binary problems.

In this paper we consider the indirect approach where
binary-decomposition methods enjoy several advantages over direct methods in multiclass problems.
It is much easier and simpler to learn a set of binary classifiers than to train one unique classifier
which separates all classes simultaneously \cite{LorenaAC2008air}.
For example, a digit recognition problem can be decomposed into a set of simpler sub-problems,
which can be easily solved by linear classifiers \cite{KnerrS92ieeetnn}.
Even in such a case, the performance is comparable to that obtained by a more complex classifier.
A comparison study \cite{HsuCW2002ieeetnn} observed that the direct methods, such as multiclass SVM
\cite{WestonJ99esann,CrammerK2001jmlr}, generally require more training time than binary-decomposition methods.
It was also observed in \cite{HsuCW2002ieeetnn} that APs-based decomposition methods show
higher predictive accuracy than the multiclass SVM for most of cases.
Moreover, in the case of binary-decomposition methods, binary classifiers can be independently trained on different processors,
which is well suited to parallel processing in the training phase.

Reducing multiclass problems to multiple binary problems can be viewed as {\em encoding},
since binary codewords are assigned to class labels.
Several encoding methods are widely used, including APs, OVA, and
error correcting output code (ECOC) \cite{DietterichTG95jair}.
Aggregation of binary classifiers involves combining prediction results determined by binary classifiers
into a final answer to the multiclass problem.
Aggregation methods can be categorized into two types: {\em hard decoding} and  {\em probabilistic decoding}.

In hard decoding, one seeks a codeword which best matches binary predictions, to determine a most probable label.
Hamming distance is often used as a discrepancy measure between a codeword
and binary predictions, in the case where individual binary classifiers yield binary outputs.
Various loss functions (such as exponential loss and logistic loss) are considered
in the case where binary classifiers yield a score whose magnitude is a measure of confidence in
the prediction, referred to be as {\em loss-based decoding} \cite{AllweinEL2000jmlr}.
In many applications, however, class membership probabilities need to be computed,
which is not possible in the hard decoding.
For instance,  in the case of cost-sensitive decision \cite{DomingosP99kdd,MargineantuDD2002ecml,ZadroznyB2001icml},
the Bayes optimal prediction is to assign an example to the class label that has a lowest expected cost
(which is also called {\em conditional risk} \cite{Duda2001}).
To this end, one needs to correctly calculate class membership probabilities for the given data point.

In probabilistic decoding, we are given binary class membership probability
estimates (scores in [0,1]) determined by binary classifiers.
One couples these probability estimates to determine
a set of class membership probabilities for multiclass problems.
In the case of APs, Hastie and Tibshirani \cite{HastieT98as}
developed a method, {\em pairwise coupling}, in which
pairwise class membership probability estimates are combined to form
a joint probability estimates for all $K$ classes,
fitting the {\em Bradley-Terry model} \cite{BradelyRA52biometrika}
by minimizing a KL-divergence criterion.
This was extended for arbitrary code matrix
(OVA and ECOC in addition to APs) \cite{ZadroznyB2001nips,HuangTK2006jmlr},
where a generalized Bradley-Terry model \cite{HuangTK2006jmlr}
was considered to relate probability estimates
obtained by binary classifiers to class membership probability estimates.

The (generalized) Bradley-Terry model provides a natural way to relate
probability estimates computed by binary classifiers to class membership probabilities,
but there are some drawbacks.
Most of aforementioned methods based on the Bradley-Terry model treat all binary classifiers equally,
leading to the performance degradation in the presence of bad binary classifiers.
This problem is alleviated by introducing confidence weights placed on individual binary classifiers
that are optimally tuned based on training data \cite{YukinawaN2009ieeetcbb}.
However, the method in \cite{YukinawaN2009ieeetcbb} involves a huge number of parameters, $NK + M$,
where $N$ is the number of training data points and $M$ is the number of binary classifiers.
In other words, the computational complexity scales linearly with the number of training data points,
which makes the method prohibitive even for mid-scale problems.
Moreover, additional iterative optimization is required to estimate the class memberships probabilities for test data.

Takenouchi and Ishii \cite{TakenouchiT2009nc} proposed a different type of decoding method
in which misclassification in binary classifier is formulated as
a bit inversion error problem, as in information transmission theory.
The dependency between classifiers are directly modeled by Bolzmann machine and
the hard decoding problem (which can also be extended to probabilistic decoding) is formulated as
a probabilistic inference problem in Bolzmann machine. The method provides
a new viewpoint to the multiclass problems in the context of information transmission theory.
However it involves exponential-order computational complexity, due to the partition function in the Bolzmann machine,
requiring approximate inference techniques such as Monte Carlo Markov Chain (MCMC) or mean filed approximation.
It might suffer from multiple local minima and is sensitive to initial conditions.

Recently, we have developed a Bayesian aggregation method \cite{ParkSH2010icdm} for probabilistic decoding.
In contrast to most of existing probabilistic decoding methods where the Bradley-Terry model was used to
relate binary probability estimates to class membership probabilities,
we directly modeled class membership probabilities as {\em softmax function}
whose input argument is a linear combination of discrepancies induced by binary classifiers.
In this way, aggregation weights are the only parameters to be tuned ($M$), while the existing method
\cite{YukinawaN2009ieeetcbb} scales linearly with the number of samples ($NK + M$).
Based on the likelihood modeled by the softmax function and the appropriate prior on the aggregation weights,
we formulated the problem of estimating aggregation weights
as variational logistic regression in which predictive distribution yielded class membership probabilities.
In such a case, regularization parameter was learned in Bayesian framework and
over-fitting was alleviated, compared to maximum likelihood methods.

There are two computational issues in the Bayesian aggregation framework:
(1) the solution suffers from local minima;
(2) the evaluation of class membership probabilities for data instances
requires additional computations (through variational optimization).
To solve these problems, one can consider the maximum likelihood estimation
instead of full Bayesian learning \cite{ParkSH2010icdm},
in which class membership probabilities can be easily computed by evaluating
the softmax function with the learned aggregation weights.
In our previous work \cite{ParkSH2011icassp}, we proposed
the $\ell_1$ norm regularized maximum likelihood method
to determine the optimal aggregation weights, which is a convex problem.
We then convert the convex optimization problem to an equivalent {\em geometric programming}
in order to make use of an off-the-shelf optimization toolbox.
With this approach, a global solution is determined and class membership probabilities
can be easily evaluated without additional optimizations.
However, our previous method \cite{ParkSH2011icassp} still has several limitations:
(1) the optimization problem can be directly solved by
the standard convex optimization algorithms without transforming it to geometric programming;
(2) it only allows $\ell_1$ norm regularization.
In contrast to \cite{ParkSH2011icassp} where the problem was converted to geometric programming,
we directly solve the optimization problem using {\em primal-dual interior point} method that is an efficient solver for convex optimization problems,
which allows us to use various types of regularization.
Especially, when $\ell_2$ norm regularization considered,
we can provide an interesting connection of our method to the large margin formulation.
The main contribution of this paper is summarized below.
\begin{itemize}
\item
Our method is more computationally efficient than the existing probabilistic decoding methods.
In our formulation, the aggregation weights are the only parameters to be tuned ($M$), while the existing method
\cite{YukinawaN2009ieeetcbb} scales linearly with the number of samples ($NK + M$).
\item
We formulate the regularized maximum likelihood estimation
as a convex optimization, so a global solution is found.
We use the primal-dual interior point method to solve this optimization problem.
\item
Connections of our method to large margin classifiers are presented, showing that
the large margin formulation can be considered as a limiting case of our convex formulation.
Moreover, we present data-dependent generalization error bound, based on margins and  Rademacher complexity,
extending existing work on binary problems \cite{BartlettPL2004jmlr} to our multiclass problems
which are solved by aggregating binary solutions.
\end{itemize}

The rest of this paper is organized as follows.
The next section describes notations and preliminaries which are needed to explain our method.
Section \ref{sec:main} provides the main contribution, in which we describe our model and show
how an optimal aggregation of binary classifiers is formulated as a convex optimization,
which is solved by the primal-dual interior point method.
Connections to large margin classifiers and generalization error bound are described in Section \ref{sec:connections}.
Experiments on synthetic and real-world data sets are provided in Section \ref{sec:simulations},
demonstrating that our method outperforms existing aggregation methods in terms of
the classification accuracy and the quality of class membership probabilities.
Finally conclusions are drawn in Section \ref{sec:conclusions}.
In addition, the appendix provides details about the proof of propositions in Section \ref{sec:connections}.

\section{Preliminaries}
\label{sec:preliminaries}

We are given $N$ training examples $\{(\bx_i, y_i)\}_{i=1}^{N}$,
where $\bx_i \in \calX \subset\Real^D$ are data vectors and $y_i \in \calY=\{1,\ldots,K\}$  ($K \geq 3$)
are class labels associated with $\bx_i$.
Multiclass prediction involves estimating the {\em class membership probabilities} of $\bx_i$,
\be
P_{k,i} \triangleq P( y_i = k \,|\, \bx_i),
\ee
for $k=1,\ldots,K,$ and $i=1,\ldots,N.$
A class label for $\bx_i$ is determined by
\bee
\widehat{y}_i = \argmax_{k} P_{k,i}.
\eee
We denote by $\bp_i = [P_{1,i}, \ldots, P_{K,i}]^{\top} \in \Real^{K}$
the class membership probability vector for data point $\bx_i$.
We also define the data matrix as $\bX = [\bx_1, \ldots, \bx_N]$ and
the class label vector as $\by = [y_1,...,y_N]^{\top}$.

Multiclass problems are decomposed into a set of binary problems that are solved by individual binary classifiers.
Such decomposition can be viewed as {\em encoding} and various methods are widely used:
\begin{itemize}
\item
OVA involves a set of $K$ binary functions, each of which discriminates one class from
the other classes.
\item
APs learns a set of $\frac{K(K-1)}{2}$ binary classifiers,
each of which distinguishes each pair of classes.
\item
ECOC assigns a binary codeword to each class such that
Hamming distances between codewords are maximized (to increase the separability) and the length
of codewords determines the number of binary functions to be learned.
\end{itemize}

Aforementioned encoding methods yield a {\em code matrix} $\bC =[C_{j,k}]\in \Real^{M \times K}$
where $M$ is the number of binary classifiers involved and $K$ is the number of class labels.
For instance, Table \ref{tab:example_APs} shows the $3 \times 3$ code matrix for a 3-class problem
in the case of APs coding.
According to the code matrix, multiclass problem is reduced to
a set of binary problems that are solved independently.
Each column in the code matrix $\bC$, denoted by $\bc_i$, corresponds to {\em codeword},
while each row defines a binary problem to be solved by a binary classifier ($BC_i$).
For instance, $BC_2$ discriminates class 2 from class 3, while samples in class 1 is not used.

\begin{table}[!ht]
\caption{code matrix in the case of APs for 3-class problem is shown,
where $BC_i$ denote binary classifiers,
1 and 0 represent positive and the negative class labels, and $\triangle$ indicates
unused class label (don't care terms).}
\begin{center}
\label{tab:example_APs}
\vspace*{.1in}
\begin{tabular}{|c|ccc|}
  \hline
                & class 1   & class 2 &  class 3\\
  \hline
        $BC_1$ &  1       &   0   &   $\triangle$       \\
        $BC_2$ &  $\triangle$       &   1        &   0 \\
        $BC_3$ &  1 &   $\triangle$        &   0       \\
  \hline
\end{tabular}
\end{center}
\end{table}

Given the code matrix $\bC$, the $j$th binary classifier is trained using examples
$\left\{ \left(\bx_i, C_{j,y_i} \right) \right\}$, where binary values of target $C_{j,y_i}$,
associated with data $\bx_i$, are determined by the code matrix.
For instance, in the case of the 2nd binary classifier in Table \ref{tab:example_APs},
the binary target value for $\bx_i$ is $C_{2,2}=1$ when $\bx_i$ belongs to 'class 2'
and is $C_{2,3}=0$ if $\bx_i$ belongs to 'class 3'.

We assume that each binary classifier yields a probabilistic prediction,
the value of which ranges between 0 and 1.
For example, we can use probabilistic SVM \cite{PlattJC99lmc}.
We denote by  $Q_{j,i}$ the probabilistic prediction by binary classifier $j$ for the class label of $\bx_i$:
\be
    Q_{j,i} \triangleq P(C_{j,y_{i}}=1 \,|\, \bx_i),
\ee
for $j=1,\ldots,M,$ and $i=1,\ldots,N.$
We denote by $\bq_i = [Q_{1,i}, \ldots, Q_{M,i}]^{\top} \in \Real^{M}$
the probability estimates computed by $M$ binary classifiers for data point $\bx_i$.
In the paper, our goal is to estimate class membership probabilities $\bp_i$ using a
collection of binary classifiers' probability estimates, $\bq_i$.

\section{Convex Optimization for Binary Classifier Aggregation}
\label{sec:main}

In this section we present our main contribution, {\em convex formulation},
where an optimal aggregation of binary classifiers into a final answer to multiclass problems
is formulated as a convex optimization, which is solved by the primal-dual interior point method.
We make use of the softmax function to relate class membership probabilities with binary probability estimates.
The softmax model takes a conic combination the discrepancies between codewords
and the probability estimates of binary classifiers, as input arguments, to represent class membership probabilities.
This approach provides a simple model to evaluate class membership probabilities,
compared to the (generalized) Bradley-Terry model-based method \cite{YukinawaN2009ieeetcbb}.

\subsection{Convex Formulation}
\label{subsec:convex}

Given probabilistic predictions of binary classifiers, $\bq_i$, for data  point $\bx_i$,
we evaluate which codeword $\bc_k$ is closest to $\bq_i$ in the sense of a pre-specified discrepancy measure,
in order to guess a class label for $\bx_i$.
To this end, we define the discrepancy $\rho(\bc_k, \bq_i, \bw)$ as a conic combination of
errors induced by $M$ binary classifiers:
\be
    \label{eq:rho}
    \rho(\bc_k, \bq_i, \bw) = \sum_{j=1}^{M} w_j \, d(C_{j,k}, Q_{j,i}),
\ee
where
\be
d(C_{j,k}, Q_{j,i})
 = -C_{j,k} \log Q_{j,i} - (1-C_{j,k}) \log (1-Q_{j,i}),
\ee
is the {\em cross-entropy} error function for two classes where the model probability for membership of
one class is $Q_{j,i}$ and the corresponding true probability is $C_{j,k}$, while the model probability for
membership of the other class is $1-Q_{j,i}$ and the corresponding true probability is $1-C_{j,k}$.
Coefficients $w_j \geq 0$ for $j=1,\ldots,M$ are {\em aggregation weights}.
In the case of $C_{j,k}=\triangle$, we do not care what a probability estimate of
the corresponding binary classifier yields, so we set $d(\triangle, Q_{j,i}) = 0$.
Our method to be described below is not restricted to the case of cross-entropy error function.
For any proper loss function, it holds.
For instance, we can also choose the exponential loss function that was used in loss-based decoding \cite{AllweinEL2000jmlr}
\be
\label{eq:exp_loss}
d_e(C_{j,k}, Q_{j,i}) = \exp\left\{ -\widetilde{C}_{j,k} (Q_{j,i}-1/2) \right\},
\ee
where $\widetilde{C}_{j,k}$=1,-1, or 0, when $C_{j,k}$=1, 0, or $\triangle$, respectively.

We define {\em aggregation weight vector} as $\bw =[w_1, \ldots, w_M]^{\top} \in \Real^{M}$.
Given data point $\bx_i$ and the probabilistic predictions $\bq_i$ determined by $M$ binary classifiers,
we model {\em class membership probability} using the softmax function that takes the form:
\be
\label{eq:soft_max}
    P(y_i = k \,|\, \bw,\bx_i)
        = \frac{\exp \left\{-\rho(\bc_k, \bq_i, \bw) \right\}}
          {\sum_{j=1}^K \exp \left\{ -\rho(\bc_j, \bq_i, \bw) \right\}},
\ee
where $\rho(\bc_k, \bq_i,\bw)$ is given in (\ref{eq:rho}).
The index $k$ yielding the smallest $\rho(\bc_k, \bq_i,\bw)$ leads to the highest class membership probability.
The prediction based on the loss-based decoding \cite{AllweinEL2000jmlr} is a special case of our model.
Fixing aggregation weights with $w_1=\cdots=w_M=1/M$,  the prediction
$\widehat{y}_i = \argmax_{k} P(y_i = k \,|\, \bw, \bx_i)$ under the model (\ref{eq:soft_max})
with the exponential loss function (\ref{eq:exp_loss}) leads to the results determined by the loss-based decoding.
In contrast, as will be explained below, we attempt to optimally tune aggregation weights using a convex optimization.

We re-arrange the class membership model probability (\ref{eq:soft_max}) by
multiplying its numerator and denominator by $\exp \left\{ \rho(\bc_k, \bq_i, \bw) \right\}$:
\be
\label{eq:sum_exp}
P(y_i = k \,|\, \bw,\bx_i)
    & = & \frac{1}{\sum_{j=1}^{K} \exp \left\{\bw^{\top} \bvarphi_i^{j,k}  \right\} },
\ee
where $\bvarphi_i^{j,k} \in \Real^{M}$ are $M$-dimensional vectors, the $l$th entry of which is given by
\be
\label{eq:varphi}
[\bvarphi_i^{j,k}]_l = d(C_{l,k},Q_{l,i}) - d(C_{l,j},Q_{l,i}).
\ee
That is, $\bvarphi_i^{j,k}$ contains differences between two discrepancies, each of which
is induced when $\bq_i$ is compared to codewords $\bc_k$ and $\bc_j$, respectively.
With these relations (\ref{eq:sum_exp}) and (\ref{eq:varphi}), we write the likelihood as
\be
\label{eq:lik_1}
\lefteqn{ p(\by \,|\, \bw,\bX) } \nonumber \\
&=& \prod_{i=1}^N \prod_{k=1}^K
\left( \frac{1}{\sum_{j=1}^{K} \exp \left\{ \bw^{\top} \bvarphi_i^{j,k}  \right\}}
\right)^{\delta(k,y_i)},
\ee
where $\delta(k,j)$ is the Kronecker delta which equals 1 if $j=k$ and otherwise 0.

We impose $\ell_2$ norm regularization on the aggregation weight vector $\bw$
and consider the negative log-likelihood, leading to the following minimization problem:
\be
    \label{eq:convex_opt}
    & & \mbox{minimize} \quad   f_0(\bw) \nonumber\\
    & & \mbox{subject to} \quad w_j \geq 0, \quad j=1, \ldots, M,
\ee
where
\bee
    \label{eq:objective_func}
    f_0(\bw) & \triangleq & -\frac{1}{N}\log p(\by \,|\, \bw,\bX)
            + \frac{\lambda}{2}\| \bw \|_2^2 \nonumber\\
        & = & \frac{1}{N}\sum_{i=1}^{N}\log \left( \sum_{j=1}^{K}
             \exp\left\{ \bw^{\top}\bvarphi_i^{j,y_i} \right\} \right)
             + \frac{\lambda}{2}\sum_{j=1}^M w_j^2,
\eee
where $\lambda>0$ is a regularization parameter.
Note that the term $\log \left( \sum_{j=1}^{K}\exp\left\{ \bw^{\top}\bvarphi_i^{j,y_i} \right\} \right)$
associated with data point $\bx_i$ in the objective function (\ref{eq:objective_func})
is called {\em log-sum-exp} which is a well known convex function used for
geometric programming \cite{BoydS2004book,BoydS2007oe}.
The objective function and constraints $w_j \geq 0$ for $j=1,\ldots,M$ are convex
in our formulation (\ref{eq:convex_opt}), so we apply a convex optimization method to
determine optimal aggregation weights $\bw$.

Note that maximum likelihood estimation is often interpreted as the minimization of
Kullback-Leibler (KL) divergence between oracle and model.
We define the true class label matrix as $\bT=[T_{k,i}] \in \Real^{K \times N}$, where
each column vector $\bt_i$ follows the 1-of-$K$ encoding to represent true class label
for $\bx_i$, such that only element associated with $y_i$ is 1 and all remaining elements equal 0.
We denote by $\bp^{w}_i \in \Real^{K}$ the $K$-dimensional class membership model probability vector
given the aggregation weight vector $\bw$, in which the $k$th element is $p(y_i = k \,|\, \bw, \bx_i)$
in (\ref{eq:soft_max}).
With these definitions, we write the KL-divergence between the oracle and the model as
\be
\label{eq:KL_dist}
\sum_{i=1}^N \mbox{KL}\Big[ \bt_{i} \,\|\, \bp^{w}_{i} \Big]
    & = & \sum_{i=1}^N \sum_{k=1}^K T_{k,i}\log \frac{T_{k,i}}{p(y_i = k \,|\, \bw, \bx_i)} \nonumber\\
    & = & \sum_{i=1}^N \log \left( \sum_{k=1}^{K}
          \exp\left\{\bw^{\top} \bvarphi_i^{k,y_i}  \right\} \right).
\ee
It follows from (\ref{eq:KL_dist}) and (\ref{eq:lik_1}) that the maximization of the likelihood (\ref{eq:lik_1})
(with the regularization ignored) equals the minimization of the KL-divergence (\ref{eq:KL_dist}).
Thus the optimal aggregation weight $\bw^{\star}$ determined by the maximum likelihood estimation
enforces $P(y_i=k \,|\, \bw^\star, \bx_i)$ to be as close as to 1 for $k=y_i$ and 0 for $k \neq y_i$.
We can also easily predict class labels for test data points, by evaluating corresponding
class membership model probabilities (\ref{eq:soft_max}) using the optimal aggregation weight vector.

\subsection{Primal-Dual Interior Point Method}
\label{subsec:pdip}

We make use of the primal-dual interior point method
\cite{FiaccoAV68book,KarmarkarN84Comb,NocedalJ99,BoydS2004book}
to solve the convex optimization problem (\ref{eq:convex_opt})
to estimate the optimal aggregation weight vector $\bw^{\star}$.
The primal-dual interior point method exhibits better than linear convergence and outperforms
the standard interior point methods in most of applications such as linear, quadratic, geometric and
semidefinite programming \cite{BoydS2004book}.
We explain the primal-dual interior point method briefly in this section to make our paper self-contained
and the algorithm is outlined in Algorithm \ref{alg:pdIP}.
Readers who are familiar with the primal-dual interior point method
can skip this section and more details can be found in \cite{BoydS2004book}.

We first examine Karush-Kuhn-Tucker (KKT) optimality conditions for the problem (\ref{eq:convex_opt}).
We denote {\em dual variables} by $\bz=[z_1, \ldots, z_M]^{\top}$ ($z_j \geq 0$ for $j=1,\ldots,M$).
Then the Lagrangian is written as
\be
    \calL(\bw,\bz) \triangleq f_0(\bw) - \sum_{j=1}^M z_j w_j.
\ee
KKT optimality conditions are given by
\be
\label{eq:kkt1}
    w_j & \geq & 0, \quad j=1,...,M, \\
\label{eq:kkt2}
    z_j & \geq & 0, \quad j=1,...,M, \\
\label{eq:kkt3}
    \nabla f_0(\bw) - \bz & = & 0, \\
\label{eq:kkt4}
    z_jw_j & = & 0, \quad j=1,...,M,
\ee
where $\nabla f_0(\bw)$ represents the gradient of $f_0(\bw)$ with respect to $\bw$.
One can easily see that Slater's constraint qualification holds for the problem (\ref{eq:convex_opt}),
since any point on the positive orthant ($\bw \in \Real_{++}^M$)
could be a strictly feasible solution to the problem \cite{BoydS2004book}.
In such a case, there exist optimal primal-dual points satisfying the KKT conditions
(\ref{eq:kkt1}) - (\ref{eq:kkt4}) and the optimal duality gap is zero.
We define $\bw^\star$ and $\bz^\star$ to be optimal primal and dual points, respectively.
Then we have
\be
    \eta \triangleq f_0(\bw^\star) - g(\bz^\star) = 0,
\ee
where $g(\bz)$ is the Lagrange dual function, i.e., $g(\bz) \triangleq \inf_{\bw}\calL(\bw,\bz)$.
The primal-dual interior point method finds the optimal primal solution $\bw^\star$ and dual solution$\bz^\star$,
which satisfy the KKT conditions (\ref{eq:kkt1}) - (\ref{eq:kkt4}).

We augment the objective function $f_0(\bw)$ by a logarithmic barrier \cite{FiaccoAV68book}
such that the constrained optimization problem (\ref{eq:convex_opt}) is converted to an
unconstrained optimization:
\be
\label{eq:logbarrier}
\mbox{minimize} \quad  f_0(\bw)  - \mu \sum_{j=1}^M \log(w_j),
\ee
where $\mu$ is the {\em barrier parameter}.
The accuracy of approximation increases as the barrier parameter $\mu$ approaches zero.
The (primal-dual) interior point methods solve the barrier sub-problems for a sequence of
the barrier parameters $\{\mu\}$ that converge to 0.
The logarithmic barrier penalizes the points that are close to zero, so
the primal solution for each barrier sub-problem is strictly feasible, i.e., $\bw(\mu)\succ 0$
(where $\bw(\mu)$ represents the solution to the optimization (\ref{eq:logbarrier}) when
a fixed value of $\mu$ is given and $\succ 0$ means that each entry in vector is greater than 0)
and eventually converges to the optimal solution as $\mu$ approaches 0.

The optimality conditions for the barrier sub-problem (\ref{eq:logbarrier})
can be interpreted as the perturbed KKT conditions.
Given the barrier parameter $\mu$,
the optimality conditions for (\ref{eq:logbarrier}) are given by
\be
\label{eq:opts}
    \nabla f_0(\bw) - \mu \bw^{-1} = 0.
\ee
where $\bw^{-1}=[1/w_1,...,1/w_M]^\top\in \Real^{M}$.
Comparing (\ref{eq:opts}) and (\ref{eq:kkt3}),
we have $z_j(\mu) = \mu/ w_j(\mu)$, where $\bz(\mu)$ is
the dual solution for the barrier sub-problem when $\mu$ is given.
With $\bw \succ 0$, the optimality conditions for
the barrier sub-problem (\ref{eq:logbarrier}) are equivalently expressed as
\be
\label{eq:mkkt1}
    z_j & \geq & 0, \quad j=1,...,M, \\
\label{eq:mkkt2}
    \nabla f_0(\bw) - \bz & = & 0,  \\
\label{eq:mkkt3}
    z_jw_j & = & \mu, \quad j=1,...,M.
\ee
The main difference between these conditions and the KKT conditions (\ref{eq:kkt1}), (\ref{eq:kkt2}),
(\ref{eq:kkt3}), and (\ref{eq:kkt4})
is in complementary slackness conditions, i.e., $z_jw_j=0$ is replaced with $z_jw_j=\mu$.
Furthermore these conditions (\ref{eq:mkkt1}), (\ref{eq:mkkt2}), and (\ref{eq:mkkt3}) can explain that
$\bw(\mu)$ converges to the optimal solution as $\mu$ approaches zero:
\be
    f_0(\bw(\mu)) - p^\star &\leq& \sum_{j=1}^M z_j(\mu) w_j(\mu) \nonumber\\
    & = & \mu M,
\ee
where $p^\star$ is a dual optimum, $p^\star \triangleq \sup_{\bz} g(\bz) = g(\bz^\star)$.
We use $\hat{\eta} = \bz(\mu)^\top \bw(\mu)$ to measure
the duality gap of the barrier sub-problem with the given $\mu$.

The iterative update rule for primal-dual interior point method
is derived by approximately solving the sequence of
the perturbed KKT conditions (\ref{eq:mkkt1}), (\ref{eq:mkkt2}), and (\ref{eq:mkkt3})  using the Newton method.
Given the barrier parameter $\mu$,
the method tries to compute the Newton step at the current solutions, $\bw(\mu)$ and $\bz(\mu)$.
With abuse of notations, we denote the current primal and dual variables by $\bw$ and $\bz$ without $\mu$.
Then, it follows from  the perturbed KKT conditions (\ref{eq:mkkt1}), (\ref{eq:mkkt2}), and (\ref{eq:mkkt3})
that the residual $r_\mu(\bw, \bz)$ is defined as
\be
    r_\mu(\bw, \bz)=
        \begin{bmatrix}
            \nabla f_0(\bw) - \bz\\
            \diag(\bz)\bw - \mu \bone
          \end{bmatrix},
\ee
where $\diag(\cdot)$ takes a vector and return a diagonal matrix with entries of the vector
placed on the diagonal, and $\bone$ is the vector of all ones.
The residual is not necessary 0 at each iteration,
except in the limits as the algorithm converges \cite{BoydS2004book}.
With a first order approximation of $r_\mu(\bw, \bz)$=0,
we can obtain the Newton step by solving the following linear equations
\be
    \label{eq:linear_system}
        \begin{bmatrix}
            \nabla^2 f_0(\bw) & -\bI\\
            \diag(\bz)  &   \diag(\bw)
        \end{bmatrix}
        \begin{bmatrix}
            \Delta\bw\\
            \Delta\bz
        \end{bmatrix}
        = -\begin{bmatrix}
            \nabla f_0(\bw) - \bz\\
            \diag(\bz)\bw - \mu\bone
          \end{bmatrix},
\ee
where $\bI$ is an $M\times M$ identity matrix.
Calculating the Newton step is further simplified by
eliminating the variable $\Delta\bz$ that can be expressed as
\be
    \label{eq:deltaz}
    \Delta\bz & = & - \diag(\bw)^{-1}\diag(\bz)\Delta\bw
               - \diag(\bw)^{-1}\Big(\diag(\bz)\bw - \mu\bone\Big).
\ee
Substituting this into (\ref{eq:linear_system}),
the linear equations are simplified as
\be
    \label{eq:linear_snd}
    \bH\Delta\bw = -\bg,
\ee
where
\be
    \bH & = & \nabla^2 f_0(\bw) + \diag(\bz) [\diag(\bw)]^{-1}, \nonumber\\
    \bg & = & \nabla f_0(\bw) - \mu\bw^{-1}. \nonumber
\ee
Note that the derivations of gradient and Hessian of the objective function
(\ref{eq:convex_opt}), $\nabla f_0(\bw)$ and $\nabla^2 f_0(\bw)$,
are provided in Appendix \ref{sec:app_grad_Hessian}.
We use the preconditioned conjugate gradient (PCG) to solve the linear system (\ref{eq:linear_snd}).
Denoting by $\bP \in \Real^{M\times M}$ the pre-conditioning matrix,
the PCG algorithm yields an approximate solution within a smaller number of steps than $M$,
when $\bP^{-1/2}\bH \bP^{-1/2}$ has just a few extreme eigenvalues.
As proposed in \cite{KhoKM2007jmlr}, we construct $\bP$
as a diagonal matrix in which the diagonal entries are set to that of $\bH$.

Given the Newton steps, $\Delta\bw$ and $\Delta\bz$,
we update the primal-dual variables:
\be
    \label{eq:update_pd}
    \bw(\mu^+) = \bw + s\Delta\bw,  \mbox{ and } \bz(\mu^+) = \bz + s\Delta\bz,
\ee
where $s$ is a step length and $\mu^+$ denotes the updated barrier parameter.
The step length $s$ can be computed by a backtracking line search as described in \cite{BoydS2004book}.
It should be carefully chosen so that $\bw(\mu^+)\succ 0$ and $\bz(\mu^+)\succeq 0$ are always satisfied.
To do this, we first compute $s^{\mbox{max}}$ to ensure $\bz(\mu^+)\succeq 0$:
\be
    \label{dq:learning_step_1}
    s^{\mbox{max}} & = & \sup\{ s\in[0,1] | \bz + s \Delta\bz \succeq 0\} \nonumber\\
            & = & \min\{ 1, \min\{-z_j/\Delta z_j | \Delta z_j < 0\} \}.
\ee
Then, we start the backtracking with $s=0.99s^{\mbox{max}}$, and
multiply $s$ by $\beta$ until we have $\bw(\mu^+) \succ 0$ and
\be
    \label{dq:learning_step_2}
    \|r_{\mu}(\bw(\mu^+),\bz(\mu^+))\|_2 \leq \|r_{\mu}(\bw,\bz)\|_2 (1 - \alpha s ),
\ee
where $\alpha$ is a small constant ($\alpha=0.01$ was used in our experiments).
In order to update the barrier parameter $\mu$,
we use an adaptively strategies that determines it
according to the reduction of the duality gap as in \cite{KhoKM2007jmlr}:
\be
    \label{eq:update_mu}
    \mu^+ = \left\{ \begin{array}{ll}
        \hat{\eta}/(2M) & \mbox{if $s \geq s^{\mbox{min}}$};\\
        \mu             & \mbox{otherwise}.\end{array} \right.
\ee
where we use $s^{\mbox{min}}=0.5$.

The primal-dual interior point method to solve the convex optimization problem
(\ref{eq:convex_opt}) is summarized in Algorithm \ref{alg:pdIP}.
The most dominant operation in Algorithm \ref{alg:pdIP} is
to compute the Newton step $\bDelta \bw$ at each iteration,
which involves calculating Hessian matrix of the objective function, $\nabla^2 f_0(\bw)$,
and solving the linear system (\ref{eq:linear_snd}).
Given a set of $\{ \bvarphi_i^{j,y_i}\}$, whose construction time is $\calO(MNK)$,
we can form $\nabla^2 f_0(\bw)$ at a cost of $\calO(M^2NK)$.
For convenience, we assume that the linear system is solved by the standard matrix inversion at a cost of $\calO (M^3)$,
while the PCG algorithm generally requires less computational cost.
Thus the total cost of computing the Newton direction is $\calO(M^2NK + M^3)$,
which is the same as $\calO(M^2NK)$ when there are more training points than
binary classifiers involved into the binary decomposition for a multiclass problem.

\begin{algorithm}
  \caption{Primal-dual interior point method for convex aggregation}
  \label{alg:pdIP}
  \SetAlgoLined
  \KwData{$\bX, \by,\bC$}
  \KwResult{Optimal aggregation weights $\bw^\star$}
  \textbf{Binary classifications}\\
    Solve the set of binary classification problems:  \\
    $\quad$ obtain $\bq_i$ for $i=1,...,N$, \\
    $\quad$ compute $\{ \bvarphi_i^{k,y_i} \}_{k=1}^K$ for $i=1,...,N$ by (\ref{eq:varphi}),\\
  \textbf{Primal-dual interior point method}\\
    Initialize parameters \\
    $\quad$ set $\alpha=0.01$ and $\beta=0.5$, $\mu = (\bw^\top \bz)/2M$,\\
    $\quad$ set $w_j = 1/M, \mbox{ and } z_j = 1, \quad \mbox{for } j=1,...,M$, \\
    $\quad$ set tolerance parameters $\epsilon_{fea}=\epsilon=10^{-4}$.\\
  \Repeat{$\|r_\mu(\bw(\mu^+),\bz(\mu^+))\| \leq \epsilon_{fea}$ and $\hat{\eta}\leq\epsilon$}{
        1. Update the barrier parameter $\mu$ using (\ref{eq:update_mu}).\\
        2. Calculate the Newton steps, $\Delta\bw$ and $\Delta\bz$:\\
        $\quad$ - compute $\nabla f_0(\bw)$ and $\nabla^2 f_0(\bw)$ (see Appendix \ref{sec:app_grad_Hessian}),\\
        $\quad$ - solve $r_\mu(\bw, \bz)=0$ using by Newton method.\\
        3. Update with line search:\\
        $\quad$ - determine the learning step $s$,\\
        $\quad$ - update the primal-dual variables by (\ref{eq:update_pd}).\\
    }
\end{algorithm}

\section{Connections to Large Margin Classifiers}
\label{sec:connections}

In this section we show the connections of our convex formulation to large margin classifiers,
in which the discrepancy differences $\{\bvarphi_i^{k,y_i}\}_{k=1}^{K}$ induced by binary classifiers
(instead of training examples $\bx_i$) are inputs to a large margin classifier.
Following the work in \cite{ZhangT2001ir} where a close relation between large margin and
logistic regression formulations is shown in the case of binary classification,
we provide its multiclass extension in this section.
We emphasize that the large margin formulation can be understood as  a limiting case of our convex formulation.

We assume that we assign data point $\bx_i$ to class $\widehat{y}_i$ if
\bee
    \label{eq:predict_label}
    \widehat{y}_i = \argmin_k \rho(\bc_k, \bq_i, \bw),
\eee
where $\rho(\bc_k,\bq_i,\bw) = \sum_{j=1}^M w_j d(C_{j,k},Q_{j,i})$ is defined in (\ref{eq:rho}).
Then the misclassification error is given by
\be
    \label{eq:empirical_err}
    \calE(\bw) = \frac{1}{N}\sum_{i=1}^N \bone \big( \widehat{y}_i \neq y_i \big),
\ee
where $\bone(\pi)$ is the {\em 0-1 loss function} which equals 1 if the predicate $\pi$ is true, otherwise 0.

\begin{figure*}[htp]
\centerline{\hbox{
\epsfig{file=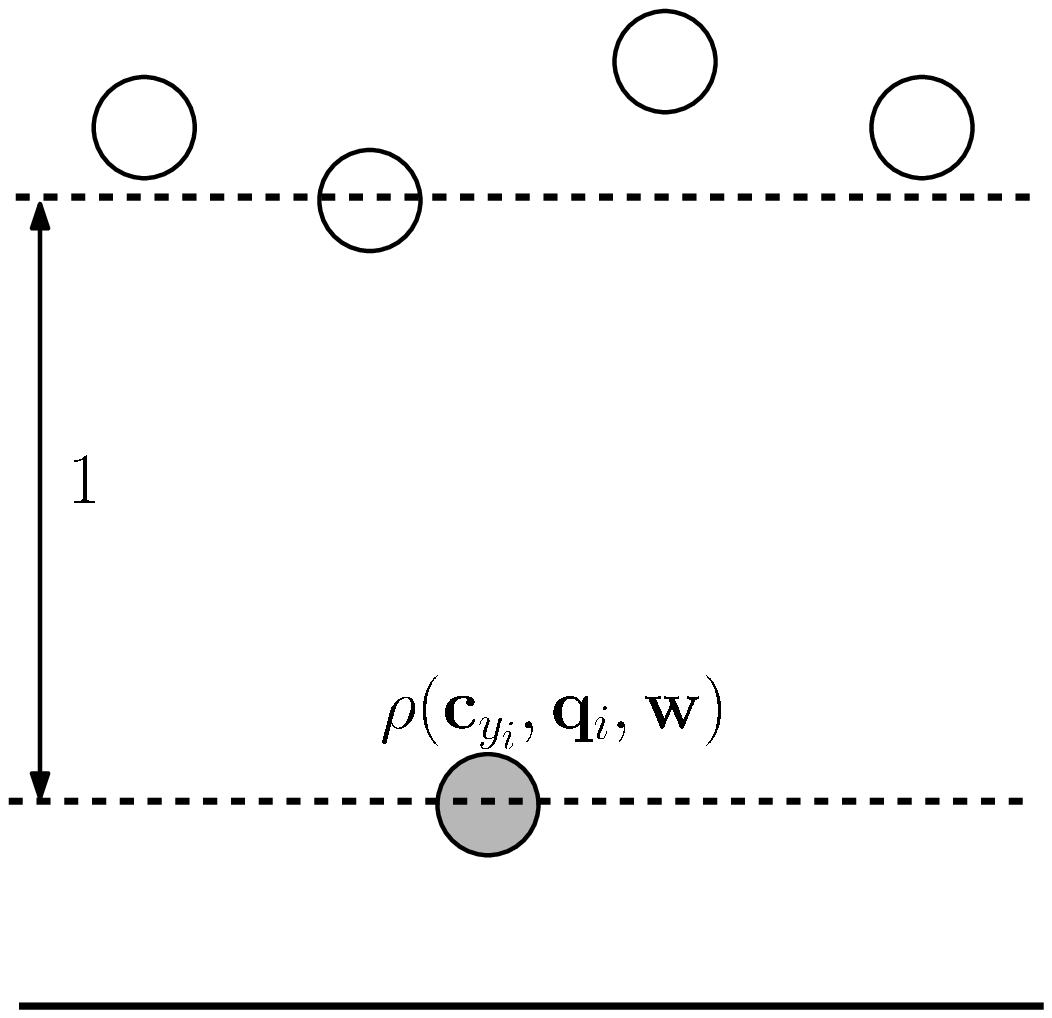,width=1.7in}
\hspace*{.3in}
\epsfig{file=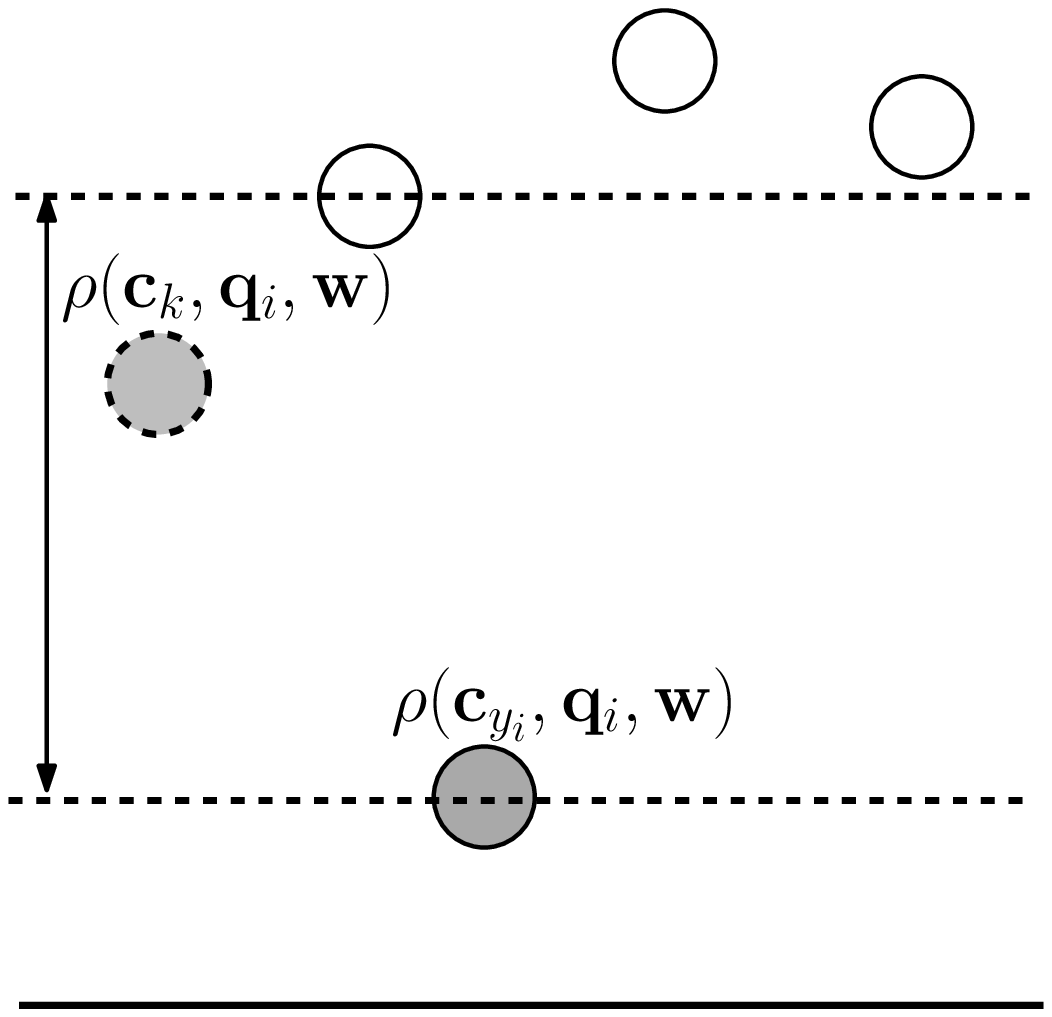,width=1.7in}
\hspace*{.3in}
\epsfig{file=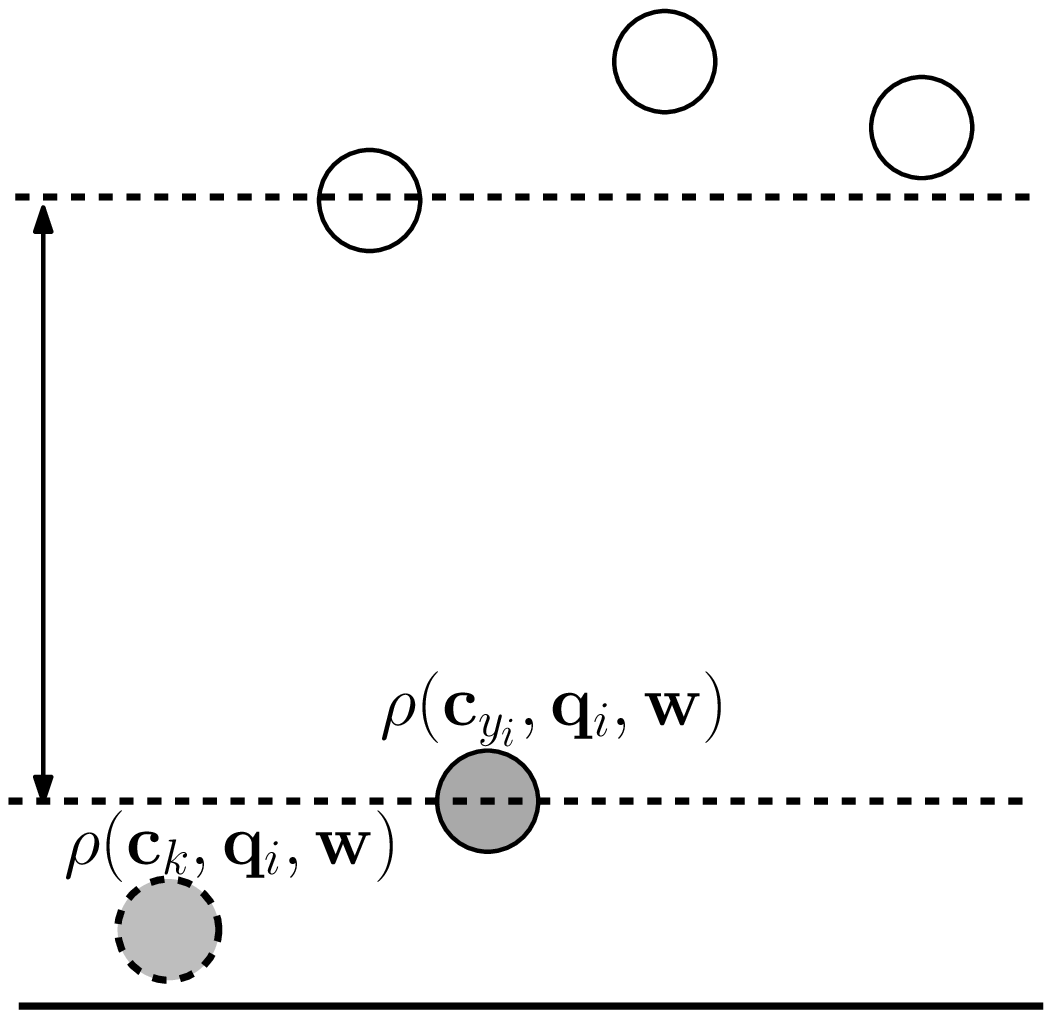,width=1.7in}}}
\vspace*{.1in}
\centerline{\hbox{
(a) \hspace*{2in} (b) \hspace*{2in} (c) }}
\caption{Pictorial illustrations for three difference cases where the margin,
$\nu_w(\bx_i,y_i)$, is: (a) greater than or equal to 1; (b) between 0 and 1; (c) less than 0.
Cases (a) and (b) yield correct predictions of labels.
In the case (c), the true label is $y_i$ but some codeword $\bc_k$ yields
the smaller discrepancy than the correct codeword $\bc_{y_i}$, leading to {\em misclassification}.
The value produced by the hinge loss function is:
(a) $h \left(\bvarphi_i^{1,y_i}, \bvarphi_i^{2,y_i}, \ldots, \bvarphi_i^{K,y_i}, \bw \right) = 0$
when $\min_{k \neq y_i}\rho(\bc_k,\bq_i,\bw) - \rho(\bc_{y_i},\bq_i,\bw) \geq 1$;
(b) $h \left(\bvarphi_i^{1,y_i}, \bvarphi_i^{2,y_i}, \ldots, \bvarphi_i^{K,y_i}, \bw \right) = 1
-\min_{k \neq y_i}\rho(\bc_k,\bq_i,\bw) + \rho(\bc_{y_i},\bq_i,\bw)$ when
$0 \leq \min_{k \neq y_i}\rho(\bc_k,\bq_i,\bw) - \rho(\bc_{y_i},\bq_i,\bw) < 1$;
(c) $h \left(\bvarphi_i^{1,y_i}, \bvarphi_i^{2,y_i}, \ldots, \bvarphi_i^{K,y_i}, \bw \right) \geq 1$
when $\min_{k \neq y_i}\rho(\bc_k,\bq_i,\bw) - \rho(\bc_{y_i},\bq_i,\bw) < 0$.}
\label{fig:hingeloss}
\end{figure*}

A direct optimization of the misclassification error (\ref{eq:empirical_err}) is not an easy task
due to the discrete nature of the 0-1 loss.
The multiclass hinge loss function, which is a convex upper bound on the 0-1 loss
$\bone \big(\widehat{y}_i \neq y_i \big)$ \cite{CrammerK2001jmlr}, was considered as a surrogate function:
\be
\label{eq:hinge_upper}
   \calE(\bw) \leq \frac{1}{N}\sum_{i=1}^N
    h \left(\bvarphi_i^{1,y_i}, \bvarphi_i^{2,y_i}, \ldots, \bvarphi_i^{K,y_i}, \bw \right),
\ee
where the hinge loss function $h \left(\bvarphi_i^{1,y_i}, \bvarphi_i^{2,y_i}, \ldots, \bvarphi_i^{K,y_i}, \bw \right)$
is given by
\be
\label{eq:hw}
\lefteqn{ h \left(\bvarphi_i^{1,y_i}, \bvarphi_i^{2,y_i}, \ldots, \bvarphi_i^{K,y_i}, \bw \right) } \nonumber \\
         & = & \max_{k \in \calY \setminus y_i} \left[ 1 -\Big(  \rho(\bc_k,\bq_i,\bw) -
         \rho(\bc_{y_i},\bq_i,\bw) \Big) \right]_+ \nonumber\\
         & = & \max_{k \in \calY} \left\{  \left(1-\delta(y_i,k) \right) + \bw^\top \bvarphi_i^{k,y_i} \right\}.
\ee
where $[a]_+=\max\{a,0\}$ and $\bvarphi_i^{y_i,y_i}=0$ is used to arrive at the second equality.
To validate the inequality (\ref{eq:hinge_upper}), we define {\em margin} as
\be
    \label{eq:margin}
    \nu_w(\bx_i,y_i) = \min_{k \neq y_{i}}\rho(\bc_k,\bq_i,\bw) - \rho(\bc_{y_i},\bq_i,\bw).
\ee
The multiclass hinge loss function (\ref{eq:hw}) yields 0
only when the margin is greater than or equal to 1.
When the margin is between 0 and 1, 
the predicted class label $\widehat{y}_i$ is still correct, i.e., $\widehat{y}_i = y_i$  but the loss
$1 -\nu_w(\bx_i,y_i)$ (which is less than 1) is produced by the hinge loss function.
The negative value of margin, where $\hat{y}_i\neq y_i$,  implies
$h \left(\bvarphi_i^{1,y_i}, \bvarphi_i^{2,y_i}, \ldots, \bvarphi_i^{K,y_i}, \bw \right) \geq 1$,
leading to (\ref{eq:hinge_upper}).
Pictorial illustration is shown in Figure \ref{fig:hingeloss}.

Thus, the problem of estimating aggregation weights can be formulated as
the large margin learning with the nonnegativity constraints on aggregation weights:
\be
    \label{eq:hard_opt}
    \mbox{minimize}_{\bw} & & f_{LM}(\bw)  \nonumber \\
    \mbox{subject to} & &  w_j \geq 0, \quad j=1,...,M,
\ee
where
\bee
 \lefteqn{f_{LM}(\bw)}  \nonumber\\
    & = & \frac{1}{N}\sum_{i=1}^N h \left(\bvarphi_i^{1,y_i}, \bvarphi_i^{2,y_i},
        \ldots, \bvarphi_i^{K,y_i}, \bw \right) + \frac{\lambda}{2} \| \bw \|_2^2.
\eee
In this setting, we seek an aggregation weight vector $\bw$
such that the empirical misclassification is minimized, while the margin is maximized.
The projected subgradient methods \cite{BertsekasDP99book} can be applied to directly solve
the primal form (\ref{eq:hard_opt}).
Note that the projected subgradient method is a first-order method which exploits the gradient only,
thus its performance much depends on the problem scaling or conditioning \cite{BoydS2007note,NocedalJ99}.
On the other hand, the primal-dual interior point method used in our aggregation method is a second-order method
where the gradient and Hessian information are exploited, so its performance is not affected by the problem scaling.
In general, (projected) subgradient methods are slower than (primal dual) interior point methods \cite{BoydS2007note}.
Thus, our convex formulation (\ref{eq:convex_opt}) benefits from (primal-dual) interior point methods,
compared to the large margin formulation (\ref{eq:hard_opt}).

We now show a close connection between our convex formulation (\ref{eq:convex_opt}) and
the large margin formulation (\ref{eq:hard_opt}).
To this end, we slightly modify the class membership model probability (\ref{eq:soft_max}),
introducing misclassification cost $1-\delta(y_i,j)$ and stiffness parameter $\tau > 0$:
\be
\label{eq:soft_max_Cost}
\lefteqn{ P(y_i = k \,|\, \bw, \bx_i) } \nonumber \\
    & = & \frac{\exp \left\{ \tau \Big( (1-\delta(y_i,k)) -\rho(\bc_{k}, \bq_i,\bw) \Big) \right\}}
        {\sum_{j=1}^K \exp \left\{ \tau \Big( (1-\delta(y_i,j)) -\rho(\bc_j, \bq_i,\bw) \Big) \right\}}, \nonumber\\
    & = &  \frac{1}{\sum_{j=1}^{K} \exp \left\{ \tau \Big( (\delta(y_i,k) - \delta(y_i,j)) + \bw^{\top}
    \bvarphi_i^{j,k} \Big)  \right\} }, \nonumber \\
\ee
This modification leads to the following minimization problem for estimating aggregation weights:
\be
    \label{eq:convex_opt_cost}
    & & \mbox{minimize} \quad f_{\tau}(\bw), \nonumber\\
    & & \mbox{subject to} \quad w_j \geq 0, \quad j=1, \ldots, M,
\ee
where
\bee
 \lefteqn{f_{\tau}(\bw)} \\
    & = & ~ \frac{1}{\tau N}\sum_{i=1}^{N}\log \Big( \sum_{j=1}^{K}
        \exp\Big\{ \tau \Big( (1 - \delta(y_i,j)) +  \bw^{\top}\bvarphi_i^{j,y_i} \Big)\Big\} \Big)
        + \frac{\lambda}{2}\|\bw\|_2^2.
\eee
The objective function  $f_{\tau}(\bw)$ is nothing but the negative log-likelihood defined by the modification
(\ref{eq:soft_max_Cost}), with $\ell_2$ norm regularization.
The parameter $\tau$ controls the stiffness and  $1-\delta(y_i,j)$ leads to a shift of the loss function
$f_0(\bw)$ (\ref{eq:objective_func}) by 1 when the Kronecker delta equals 0.
Proposition \ref{prop:p1} outlines that the large margin formulation can be interpreted
as a limiting case of our convex formulation (\ref{eq:convex_opt_cost}).

\begin{proposition}
\label{prop:p1}
The sequence of functions $\{f_\tau(\bw)\}$ ($\tau=1,2,...$) uniformly converges
to the objective function $f_{LM}(\bw)$ in the large margin formulation (\ref{eq:hard_opt}).
That is, given any $\epsilon > 0$, there exists a natural number $\Xi = \Xi(\epsilon)$ such that
\bee
\left| f_\tau(\bw) - f_{LM}(\bw) \right| < \epsilon,~\mbox{for $\forall$ $\tau > \Xi$
and $\forall$ $\bw \in \Real^{M}$}.
\eee
\end{proposition}
{\em Proof.} See Appendix \ref{sec:app_proof1}.

We would like to point out a few things about our convex formulations (\ref{eq:convex_opt}) and
(\ref{eq:convex_opt_cost}), and the large margin formulation (\ref{eq:hard_opt}).
\begin{itemize}
\item
A special case of (\ref{eq:convex_opt_cost}) when fixing $\tau=1$ and
neglecting the misclassification loss $1-\delta(y_i,j)$,
leads to our original convex formulation (\ref{eq:convex_opt}).
\item
In contrast to the large margin formulation (\ref{eq:hard_opt}), the convex formulation
(\ref{eq:convex_opt_cost}) allows us to calculate the gradient and Hessian, so the primal-dual interior point
method can be used to find the optimal value of $\bw$, as in the case of (\ref{eq:convex_opt}),
while the subgradient method is used to solve the large margin formulation (\ref{eq:hard_opt}).
\item
Solving (\ref{eq:convex_opt_cost}) requires a successive application of the primal-dual interior point
method, gradually increasing the value of $\tau$.
Starting from $\tau=1$, the primal-dual interior point method is used to determine the optimal $\bw$.
This optimal value of $\bw$ is used as an initial condition at the next iteration with increasing $\tau$.
\item
In our experience with extensive numerical experiments, these three formulations
(\ref{eq:convex_opt}), (\ref{eq:hard_opt}) and (\ref{eq:convex_opt_cost}) yield
similar performance in terms of classification accuracy.
However, we prefer our original convex formulation (\ref{eq:convex_opt}) to others, due to its
computational efficiency and implementation simplicity.
\end{itemize}


In addition, we present a data-dependent generalization error bound, based on
the large margin formulation (\ref{eq:hard_opt}) using the Rademacher complexity \cite{BartlettPL2004jmlr}.
Our result is an extension of existing work on binary problems \cite{BartlettPL2004jmlr} to the multiclass problems
solved by aggregating binary solutions.
To this end, we treat the margin (\ref{eq:margin}) as a decision function for multiclass problems, so that
a class of functions is given by
$\calF = \{ f: \calX \times \calY \rightarrow \Real \,|\, f(\bx,y) = \bw^{\top} \bvarphi^{\overline{k},y}_{x} \}$,
where $\bw \in \Real^{M}_+$ the aggregation weight vector
and $\overline{k} = \argmin_{k \neq y} \rho( \bc_k, \bq_x,\bw).$
Note that $\bvarphi^{\overline{k},y}_{x}$ can be considered as feature mapping, as in kernel methods.
Applying Theorem 7 in \cite{BartlettPL2004jmlr} to our problem, with the empirical Rademacher complexity,
yields the following proposition.


\begin{proposition}
\label{prop:p2}
Let $P_{x,y}$ be a probability distribution on $\calX \times \calY$, from which $(\bx,y)$ is drawn.
Given $\epsilon>0$, with probability $\geq 1 - \epsilon$ over training samples  $\{(\bx_i,y_i)\}_{i=1}^N$
drawn independently from $P_{x,y}$, for every aggregation weight vector $\bw \in \Real^{M}_+$ for $\|\bw\|_2 \leq B$,
\bee
    \label{eq:error_bound_computable}
    P ( y \neq \hat{y}) \leq 
            \frac{1}{N} \sum_{i=1}^N \phi \Big( \nu_{w}(\bx_i,y_i) \Big) + \,\, \frac{2B}{N}\Bigg(\sum_{i=1}^N \min_{k \neq y_i}
            \|\bvarphi_i^{k,y_i}\|_2^2\Bigg)^{{1}/{2}} + \sqrt{\frac{9\log (2/\epsilon)}{2N}},
\eee
where $\nu_{w}(\bx_i,y_i)$ is the margin (\ref{eq:margin}) and
$\phi(z) = \min(1, \max(0, 1 - z))$, for $z \in \Real$, is the {\em ramp loss} that is
a clipped version of hinge loss \cite{CollobertR2006jmlr}.
\end{proposition}

{\em Proof.} See Appendix \ref{sec:app_proof2}.

In Proposition \ref{prop:p2}, the generalization error $P ( y \neq \hat{y}) $ is upper-bounded by a sum of
three terms, each of which is the average empirical loss, the empirical Rademacher complexity of the function class $\calF$,
and a constant term depending on $\epsilon$ (confidence parameter) as well as the number of training samples $N$.
Lemma 22 in \cite{BartlettPL2004jmlr} was used to compute the empirical Rademacher complexity in our problem.
Note that the average empirical loss in Proposition \ref{prop:p2} is not convex.
Thus Replacing the ramp loss $\phi$ by the multiclass hinge loss function $h$ (\ref{eq:hw}) that
is a convex upper bound on $\phi$, with regularization, yields the large margin formulation (\ref{eq:hard_opt})
which can be solved by convex optimization.


Proposition \ref{prop:p2} theoretically supports the validity of our aggregation method,
including the large margin formulation (\ref{eq:hard_opt}) and the convex formulation (\ref{eq:convex_opt_cost}).
The aggregation weights are determined by minimizing the average multiclass hinge loss,
equivalently maximizing the margin. Thus, our aggregation method yields the lower generalization error, since
Proposition \ref{prop:p2} implies the larger the margin the lower the generalization error of classifiers.
This is also applied to our convex formulation  (\ref{eq:convex_opt}),
due to its strong connection to the formulation (\ref{eq:convex_opt_cost}).
Note that our generalization error bound is similar to the ones for boosting with loss-based decoding \cite{AllweinEL2000jmlr},
while the error bound in \cite{AllweinEL2000jmlr} is based on VC dimension
which does not depend on sample distribution in contrast to Rademacher complexity.
In addition, it also follows from Proposition \ref{prop:p2} that our aggregation method yields the lower generalization error,
compared to the loss-based decoding, because our method minimize the average multiclass hinge loss while the loss-based
decoding use fixed aggregation weights ($w_j = 1/M$ for $j=1,\ldots,M$).

\section{Experiments}
\label{sec:simulations}

We evaluated the performance of our method on several data sets in terms of
the classification accuracy and the quality of class membership probability estimates,
compared to existing multiclass classification methods.
They include two direct methods for multiclass problems,
\{multiclass SVM (M-SVM) \cite{CrammerK2001jmlr} and lasso multinomial regression (LMR) \cite{FriedmanJ2010jss}\},
and three aggregation methods, \{loss-based decoding \cite{AllweinEL2000jmlr},
GBTM in \cite{HuangTK2006jmlr} (which is a probabilistic decoding method based on the generalized Bradley-Terry models)
and WMAP \cite{YukinawaN2009ieeetcbb}.\}
We carried out numerical experiments on two synthetic and eight real-world data sets,
in order to show the usefulness and high performance of our convex aggregation method.

\subsection{Experimental Setting}
\label{sec:experimental_setting}

M-SVM \cite{CrammerK2001jmlr} aims to directly construct a K-way classifier
which separates the correct class labels from the rest of class labels
by maximizing the margin defined as $g_{y_i}(\bx_i) - \max_{k\neq y_i} g_k(\bx_i)$,
where $\{ g_{k}\}_{k=1}^K$ are classification functions for each class.
Thus the prediction for a new point $\bx_o$ is made by $\hat{y}_o = \max_{k} g_k(\bx_o)$.
In the experiments, we used a linear kernel and the regularization parameter $\lambda_{M}$,
which trades off the empirical misclassification error and the margin, was obtained
by maximizing the classification accuracy on randomly chosen validation set on a parameter space
$\lambda_{M} \in \{10^0, 10^1,...,10^6\}$.
We used a toolbox available at http://svmlight.joachims.org/svm\_multiclass.html.

LMR \cite{FriedmanJ2010jss} is a variant of generalized liner model (GML) \cite{NelderJ72jrssA}
that generalizes linear regression by allowing a linear model to be related with
the response variables (characterized by exponential family distribution)
through the {\em response function}. Especially, multinomial regression (MR) defines
a linear model which is related with the categorical response variable (class labels).
In this case, the response function turns out to be the class membership probability in multiclass problems:
\be
	\label{eq:classprob_MSVM}
	P(y = k|\bx) = \frac{\exp\{ \gamma_{k0} + \bgamma_k^\top \bx \}}{\sum_{j=1}^K \exp\{ \gamma_{j0} + \bgamma_j^\top \bx  \}}.
\ee
where parameters are $\gamma_{j0} \in \Real$, $\bgamma_j \in \Real^D$, for $j=1,...,K$.
LMR \cite{FriedmanJ2010jss} determines the parameters
by maximum likelihood with the $\ell_1$ norm (lasso) regularization.
We used a Matlab toolbox which is available at http://www-stat.stanford.edu/~tibs/glmnet-matlab/.

For binary-decomposition methods, we used three encoding schemes,
OVA, APs and ECOC, where the code matrix $\bC$ is determined as in Section \ref{sec:preliminaries}.
The most simple case is OVA encoding: the code matrix $\bC$ is set to a $K \times K$ identity matrix.
In APs encoding, we learned a set of $M = \frac{K(K-1)}{2}$ binary classifiers,
each of which distinguishes each pair of classes.
The code matrix for APs is a $M\times K$ rectangle matrix
in which each column includes only one 1 and 0.
In the case of ECOC encoding, we used two strategies to generate the code matrix $\bC$:
complete code and sparse random code \cite{AllweinEL2000jmlr}.
For $K < 8$, we used the complete code, yielding $M=2^{K-1}-1$ binary classifiers
and generating a binary code matrix without don't care terms ($\triangle$).
For $K \geq 8$, we generated a spare random code matrix
as in \cite{AllweinEL2000jmlr}, in which $M=\lceil 15\log_2 K \rceil$,
and entries of the code matrix are chosen as $\triangle$ with probability $1/2$
and $0$ or $1$ with probability $1/4$ for each.
To increase the separability between codewords, Hamming distance $\kappa$ between
each pair of columns in $\bC$ should be large.
We selected the matrix with a maximum $\kappa$ by generating 20,000 random matrices
and ensuring that each column has at least one 0 and one 1.

We used two linear SVMs to implement the base binary classifier,
LibSVM and Liblinear \cite{FanRE2008jmlr}.
In fact, the loss-based decoding and our method do not require that
the binary classifier yields probability estimates.
However, for fair comparison with GBTM and WMAP which are based on
the probability estimates of binary classifiers,
we converted the score obtained by SVM into the probability.
In the case of LibSVM, Platt's sigmoid model \cite{PlattJC99lmc} is used
to calculate the binary class membership probability:
\be
\label{eq:Platt_SVM}
Q_{j,i} = \frac{1}{1 + \exp \left\{- Ag_{j}(\bx_i) + B \right\}},
\ee
where $g_{j}$ is the function learned by the $j$th SVM and
$A,B \in \Real$ are parameters are tuned by
the regularized maximum likelihood framework \cite{PlattJC99lmc,LinHT2007mlj}.
Note that, in the case of Liblinear the binary class membership probabilities
are directly calculated by $\ell_2$ norm regularized logistic regression.
For linear kernel case, the regularization parameter $\lambda_B$, only user parameter to be set, is obtained
by maximizing the classification accuracy on randomly chosen validation set on a parameter space $[2^{-3},2^{-2},...,2^3,2^4]$.

As mentioned in Section \ref{subsec:convex}, the loss-based decoding \cite{AllweinEL2000jmlr} can be implemented
as a special case of our aggregation method, where the aggregation weights
are set to $\widetilde{w}_1 = ... = \widetilde{w}_M = 1/M$.
In this setting, the method can be easily extended to probabilistic decoding.
The class membership probability of the instance $\bx_i$
in the loss-based decoding is given by
\be
    P(y_i = k | \bx, \widetilde{\bw}) = \frac{\exp\{ -\rho_e (\bc_k,\bq_i,\widetilde{\bw}) \} }
            {\sum_{j=1}^K \exp\{-\rho_e (\bc_j,\bq_i,\widetilde{\bw}) \}},
\ee
where $\rho_e (\bc_k,\bq,\widetilde{\bw}) = \frac{1}{M} \sum_{j=1}^M d_e(C_{j,k}, Q_{j,i})$
and $d_e$ is the exponential loss defined in (\ref{eq:exp_loss}).

WMAP \cite{YukinawaN2009ieeetcbb} is an existing optimal aggregation method,
which also tunes aggregation weights based on training data.
The method is also based on the generalized Bradley-Terry models to
connect class membership probabilities to the probability estimates obtained by binary classifiers.
The aggregation weights are assigned to each classifier an learned by
maximizing the objective function which represents the concordance
between the class membership probability estimates and the target labels \cite{YukinawaN2009ieeetcbb}.
Note that, since the aggregation weights are indirectly related with the objective function,
the calculation of the gradient of the objective function with respect to the aggregation weights is tricky
and involves the optimization of class membership probabilities for whole training data.
It usually takes too much of time to compute the gradient at each iteration, so WMAP is prohibitive even for mid-scale problems.
In the experiments, we estimated the aggregation weights using WMAP only for small datasets,
otherwise class membership probabilities for the test data were just estimated with
the fixed aggregation weights, $w_j = N_j/\sum_{j=1}^M N_j$, where
$N_j$ is the number of training points involved in the $j$th binary classification problem.
Some user parameters were manually set, choosing the values yielding the best performance after several trials were made.

GBTM is also a probabilistic decoding method based on generalized Bradely-Terry models \cite{HuangTK2006jmlr},
which can be understood as a special case of WMAP with the uniform aggregation weighs.
Similar to WMAP, class membership probabilities are computed  by minimizing KL divergence between
the generalized Bradely-Terry models and the probability estimates obtained by binary classifiers.
However, the method does not includes the learning procedure of aggregation weights:
it only provides the fixed-point type update rule for computing class membership probabilities for test points.
We implemented the method according to {\em Algorithm 2} in \cite{HuangTK2006jmlr}.
Both GBTM and WMAP were implemented in Matlab.

\begin{figure}[htp]
\centering
\centerline{\hbox{
\epsfig{file=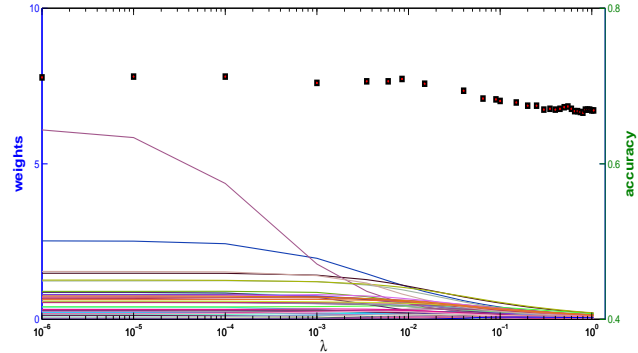,width=3.8in,height=2.0in}}}
\caption{Regularization path versus the regularization parameter $\lambda$
for vowel dataset with ECOC encoding, where the square indicates
the training accuracy at each value of $\lambda$.}
\label{fig:regularization_path}
\end{figure}

For our method, we need to determine the regularization parameter $\lambda$ in (\ref{eq:convex_opt}).
To do this, we investigated the regularization path, which explains
how the value of $\lambda$ affects the optimal solution of aggregation weights.
For example, we examined 'Vowel' dataset from UCI repository \cite{uciml2010},
which contains 11 classes and about 1000 examples.
In this case, we used LibSVM with linear kernel for the base binary classier
and ECOC (sparse random code) encoding for binary-decomposition,
in which the number of classifiers is $\lceil 15\log_2 K \rceil=52$, so is the dimension of $\bw$.
The regularization path to this problem is shown in Figure \ref{fig:regularization_path}.
We also reported the training classification accuracy of this dataset:
the square in the figure indicates the classification accuracy for the training data at each value of $\lambda$.
The method showed reasonable performance for $\lambda \leq 10^{-2}$.
With extensive numerical experiments, we found that our method shows
the stable performance for the wide range of the value of $\lambda$. For simplicity,
we set $\lambda=10^{-4}$ for all experiments.
The primal-dual interior point algorithm in Table \ref{alg:pdIP} was implemented in Matlab.
All experiments were run on Intel i7 quad-core 2.67GHz cpu with 12GB main memory.

\subsection{Synthetic Data}
\label{sec:num_results}

In this subsection, we show that how our method improves the overall classification accuracy
of the loss-based decoding by adapting the aggregation weights on the given dataset.
We first considered a 3 class problem, in which each class includes $100$ training examples.
The APs encoding was used for the binary-decomposition, so three binary classification problems
were defined based on the code matrix $\bC$ which equals Table \ref{tab:example_APs}.
We assumed that one of three classifiers fails to correctly separate the given pair of classes.
To realize this assumption, we directly generated the probability estimates for three binary classifiers,
i.e., $\{Q_{j,i}\}$ for $i=1,...,300$ and $j=1,...,3$.
Denote $\calI_k$ by a set of indexes of the training examples with class label $k$.
We assumed that the first and second classifiers are well designed for their purposes,
but, the third classifier, $BC_3$, was designed to fail to achieve its goal.
Thus, $\{Q_{j,i}\}$ were generated as follows
\begin{itemize}
\item from $BC_1$ (the classifier that separates classes 1 and 2):  \\
$Q_{1,i} = 0.9 + 0.1u_{1,i}$ for $i \in \calI_1$, $Q_{1,i} = 0.1 + 0.1u_{1,i}$ for $i\in \calI_2$
\item from $BC_2$ (the classifier that  separates classes 1 and 3): \\
$Q_{2,i} = 0.6 + 0.4u_{2,i}$ for $i \in \calI_1$, $Q_{2,i} = 0.1 + 0.1u_{2,i}$ for $i\in \calI_3$
\item from $BC_3$ (the classifier that separates classes 2 and 3): \\
$Q_{3,i} = 0.5 + 0.5r(v_{3,i})u_{3,i}$ for $i \in \calI_2$, $Q_{3,i} = 0.5 + 0.5r(v_{3,1})u_{3,i}$ for $i\in \calI_3$.
\end{itemize}
Here, $u_{j,i}$ and $v_{j,i}$ were generated from the uniform distribution,
$u_{j,i},v_{j,i} \sim \calU_{[0 \,\, 1]}$ and
$r(a)$ is a binary function that is 1 if $a>0.5$, otherwise -1.
For each binary classifier, the probabilistic estimates for the training examples
associated with unused class label $\triangle$ were assumed to be generated from the uniform distribution.
For example, in the case of $BC_1$, $Q_{1,i} = u_{1,i}$ for $i \in \calI_3$, where $u_{1,i}\sim \calU_{[0 \,\, 1]}$.
Note that the classifier $BC_3$ totally fails to separate classes 2 and 3.

The loss-based decoding method gives undesirable classification results due to the incorrect classifier, $BC_3$.
In this case, the training classification accuracy is $0.780$, and the method often misclassify classes 2 and 3.
We can check this point from the confusion matrix of the loss-based decoding method on this dataset,
shown in Table \ref{tab:confusion_loss}.

\begin{table}[!ht]
\begin{center}
\caption{Confusion matrix of the loss-based decoding method for 3 class problem.}
\label{tab:confusion_loss}
\begin{tabular}{|cc|ccc|}
  \hline
                        &                 & \multicolumn{3}{c|}{Predict}\\
                        &                 & class 1   & class 2   &   class 3  \\
  \hline
  \multirow {3}*{True}  &   class 1      &  100       &    0      &  0    \\
                        &   class 2      &   2        &    67     &  31   \\
                        &   class 3      &   5        &    28     &  67    \\

  \hline
\end{tabular}
\end{center}
\end{table}

\begin{figure}[htp]
\centering
\centerline{\hbox{
\epsfig{file=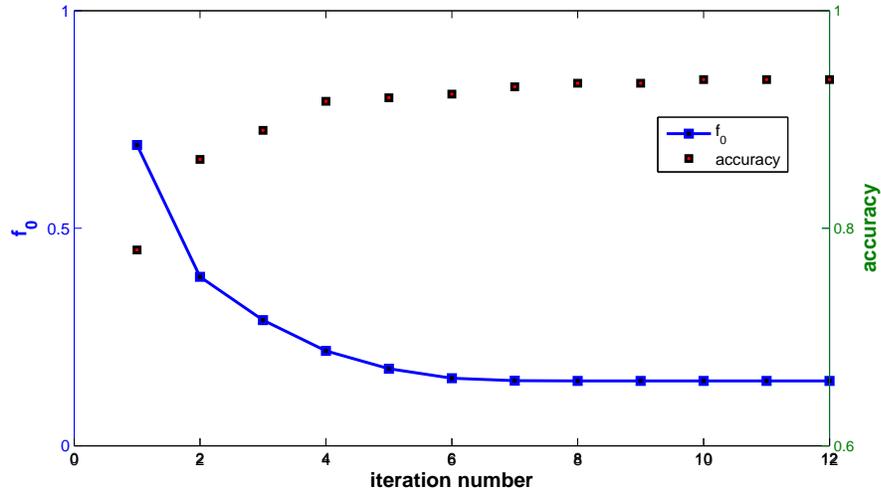,width=5.1in,height=2.8in}}}
\caption{Progress of primal-dual interior point method for 3 class dataset.
The plot shows the objective function value $f_0(\bw)$ versus the number of iteration.
The red square represents the classification accuracy evaluated at the solution of each iteration.}
\label{fig:progress_PDIP}
\end{figure}

Our method can give the better solution for this problem
by adapting the aggregation weights based on the observed data.
When the results from a certain classifiers are unreliable,
our method can remove the effect of this incorrect classifier on the overall classification accuracy
by setting the corresponding aggregation weight to as close as to zero.
We obtained the optimal aggregation weight vector $\bw^\star$ by solving the optimization problem (\ref{eq:convex_opt}).
Figure \ref{fig:progress_PDIP} shows the progress of primal-dual interior point method for this dataset.
As mentioned in Section \ref{subsec:convex}, the initial solution of the algorithm was set to
a uniform weight vector, $w_1=w_2=w_3=1/3$, in which our method produces the identical prediction to the loss-based decoding.
As the algorithm converges, the classification accuracy evaluated using $\bw(\mu)$ (the solution of each iteration) increases.
The final solution of our method, $\bw^\star$, is given by
\be
    \bw^\star = [8.3146,\quad 8.2828,\quad 0.0161]^\top.
\ee
As we expected, the aggregation weight for $BC_3$, $w_3$, becomes close to zero.
With the optimal aggregation weight vector $\bw^\star$,
the training classification accuracy is $0.940$, which is much higher than that of loss-based decoding.
The confusion matrix of our method is also given in Table \ref{tab:confusion_opt}.

\begin{table}[!ht]
\begin{center}
\caption{Confusion matrix of our aggregation method for 3 class problem.}
\label{tab:confusion_opt}
\begin{tabular}{|cc|ccc|}
  \hline
                        &                 & \multicolumn{3}{c|}{Predict}\\
                        &                 & class 1   & class 2   &   class 3  \\
  \hline
  \multirow {3}*{True}  &   class 1      &  100       &    0      &  0    \\
                        &   class 2      &  0         &    92     &  8   \\
                        &   class 3      &  0         &    10     &  90   \\

  \hline
\end{tabular}
\end{center}
\end{table}

We further compared the performance of the loss-based decoding and our method,
in the case where the number of classes is increased with the fixed number of training data.
The data instances evenly sampled from $K$ number of 2-dimensional Gaussian distributions,
the mean vectors of which are chosen as $D$ independent uniform $[0,\,\,20]$ random variables.
We allowed the overlap of classes, thus
the separation of classes might be harder as the number of classes is increased.
For each trial $1300$ samples were drawn from each Gaussian distribution,
in which $300$ samples were used for training and 1000 samples for test.
As the number of classes increase, the data points in each class became spares.
With this synthetic data, we can examine the performance of our method for sparse dataset varying with the number of classes.
We used the Liblinear (linear SVM) for the base binary classifier.

\begin{figure*}[htp]
\centering
\centerline{\hbox{
\epsfig{file=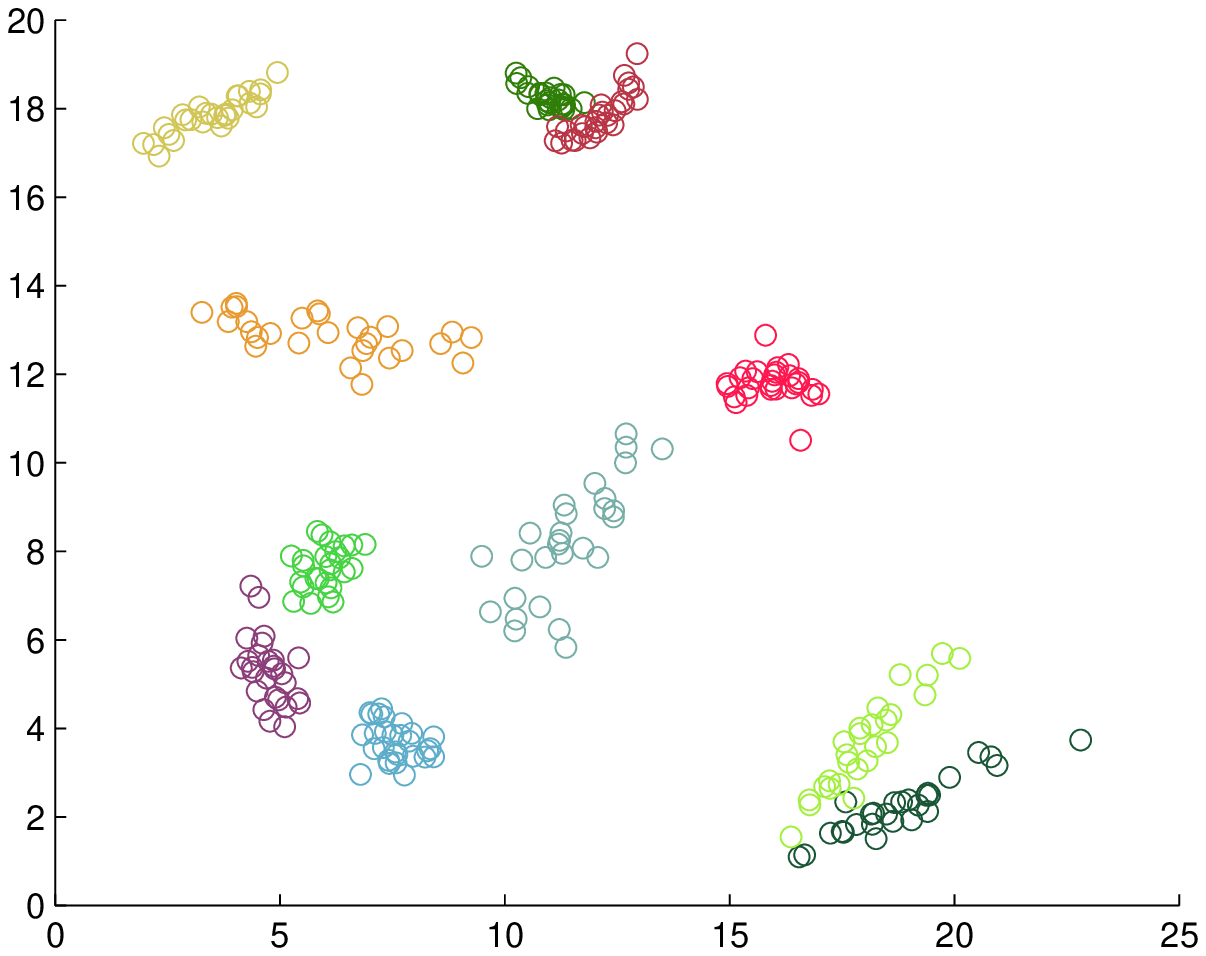,width=2.60in,height=1.65in}
\epsfig{file=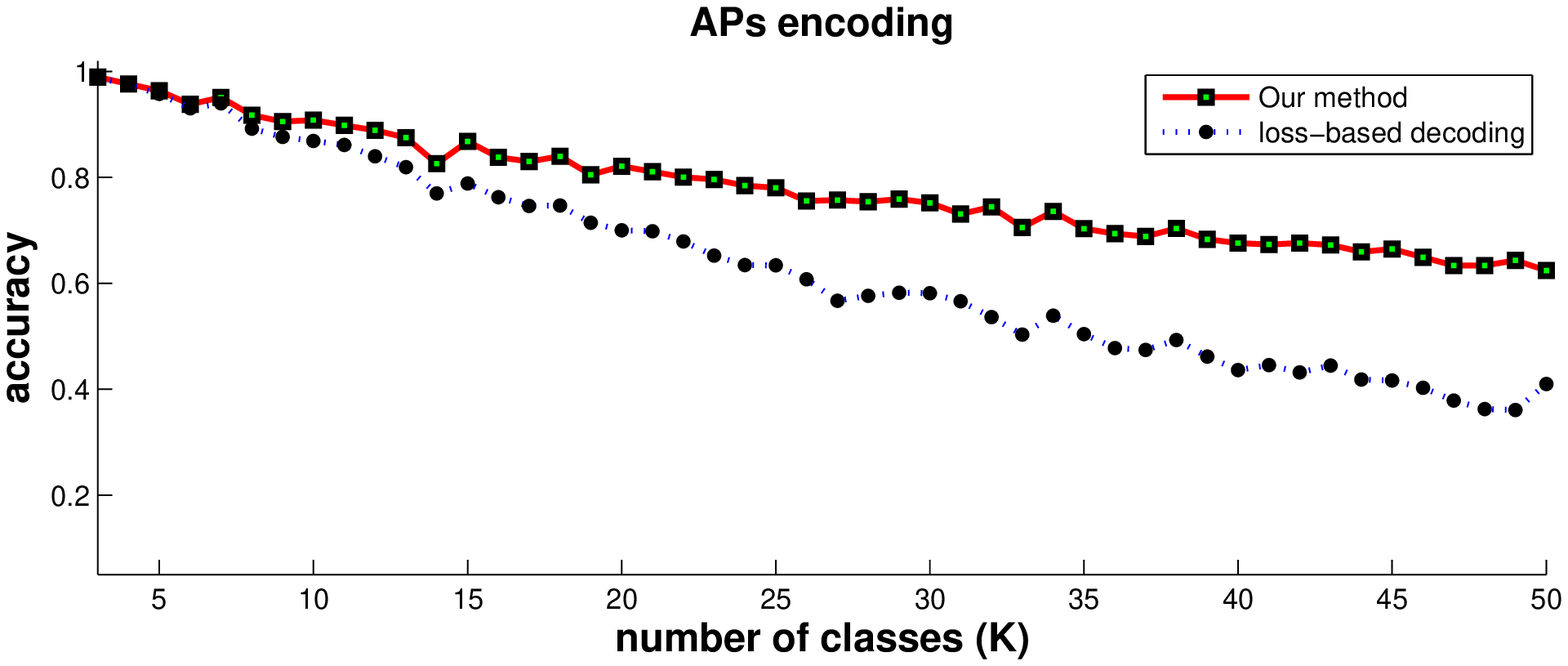,width=2.60in,height=1.65in}}}
\centerline{\hbox{(a) \hspace{6.5cm} (c)}}
\vspace{0.5cm}
\centerline{\hbox{
\epsfig{file=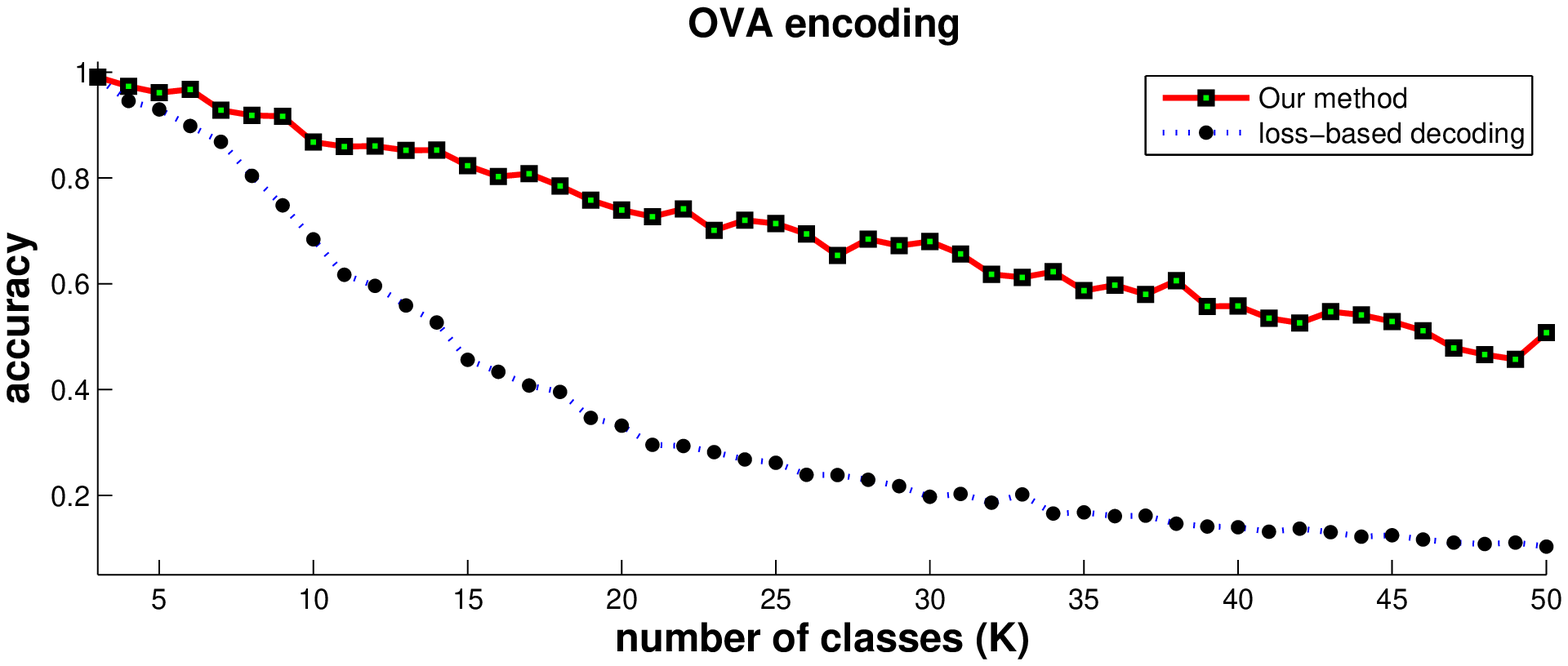,width=2.60in,height=1.65in}
\epsfig{file=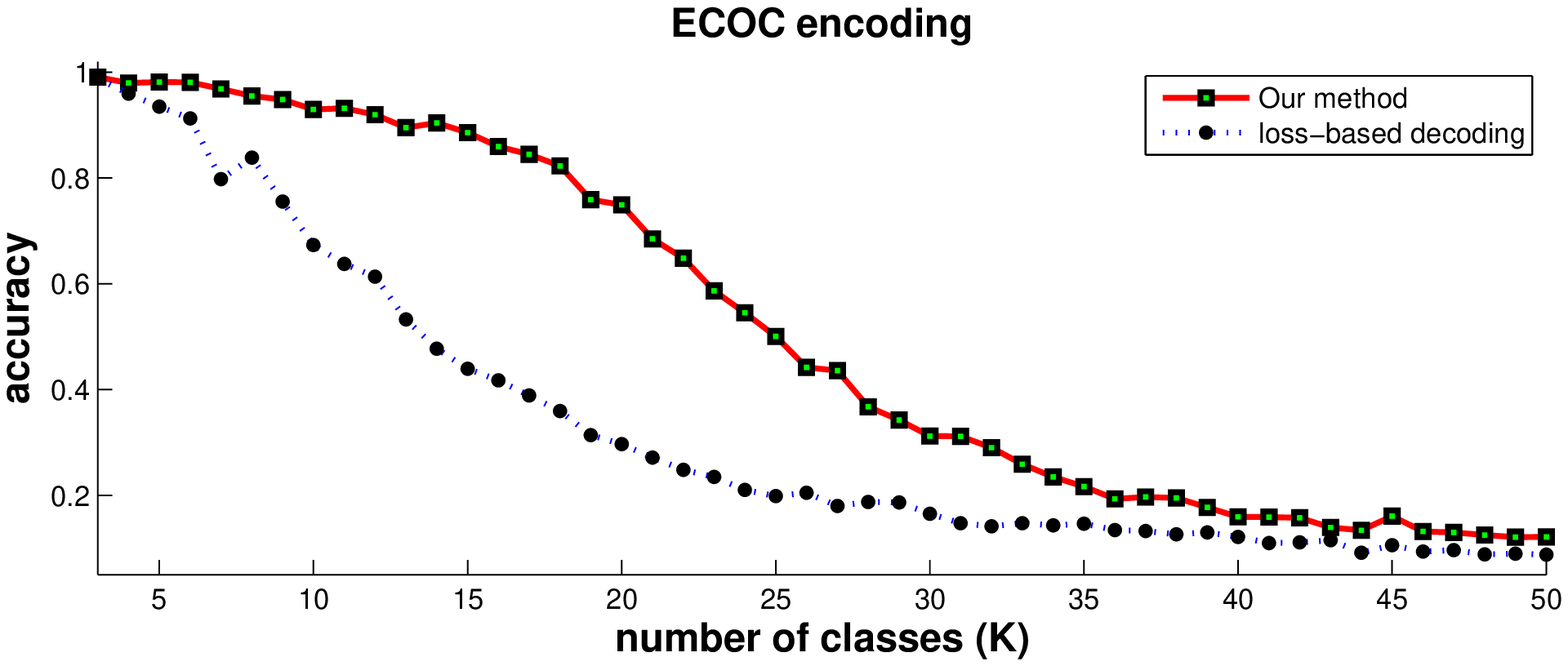,width=2.60in,height=1.65in}}}
\centerline{\hbox{(c) \hspace{6.5cm} (d)}}
\caption{Classification accuracy of the loss-based decoding and our method on a 2-dimensional synthetic dataset.
(a): One example of the synthetic dataset is shown, where the number of classes is 11 and each color represents the different class.
(b)-(d): Classification accuracy of each method is averaged over 20 independent runs,
when $K$ varies from 3 to 50, in the cases of (b) APs, (c) OVA and (d) ECOC.}
\label{fig:2d_comparision}
\end{figure*}

Figure \ref{fig:2d_comparision} (b)-(d) represent the classification accuracy averaged over 20 independent runs,
when the number of classes, $K$, varies from 3 to 50 in the cases of APs, OVA and ECOC, respectively.
Our method improves the classification accuracy of the loss-based decoding in the all cases.
In addition, we can confirm that our method well works for the case where
the number of data points in each class is relatively small compared to the number of classes.

\subsection{Real-World Data}
\label{sec:performance}

To compare the performance of each method on real-world problems,
we used two cancer datasets and six UCI data sets \cite{uciml2010}.
The cancer data sets are acute lymphoblastic leukemia (ALL) \cite{YeohEJ2002cc}
and global cancer map (GCM) \cite{RamaswamyS2001pnas},
which were used to evaluate the performance of WMAP \cite{YukinawaN2009ieeetcbb}.
The detailed descriptions for the datasets are summarized in Table \ref{tab:data_table}.
All datasets were pre-processed such that all attributes are normalized to have unit variance,
in order for attributes to reside in similar dynamic ranges.
Especially, for two cancer datasets, we only selected a subset of genes
as classification features: we chose $1,000$ genes as the input features by using a gene ranking
based on the ratio of between group- to within group sum of squares.
In addition to the cancer datasets, the input dimensionality of 'isolet' dataset was reduced
into 25 (equals to $K-1$) by Fisher linear discriminant analysis
in order to reduce the computational complexity without loss of classification performance.

\begin{table}[!ht]
\begin{center}
\caption{Data description.}
{\small
\label{tab:data_table}
\begin{tabular}{|c|ccc|}
  \hline
                       &  $\#$ samples       & $\#$ attributes     &   $\#$ classes    \\
  \hline
        ALL            &  248       &  12,558   &   6    \\
        GCM            &  198       &  16,063   &   14   \\
        glass          &  214       &    9      &   7    \\
        segmentation   &  2,310      &    19     &   7    \\
        satimage       &  6,435      &    36     &   7    \\
        pendigits      &  10,992     &    16     &   10   \\
        isolet         &  7,797      &    617    &   26    \\
        letter         &  20,000     &    16     &   26   \\
  \hline
\end{tabular}
}
\end{center}
\end{table}

In addition to classification accuracy, we evaluated mean square error (MSE)
to examine the quality of class membership probability estimates obtained by the method.
For a new test point $\bx_o$, we usually have a corresponding true label, but not class membership probabilities.
As in \cite{ZadroznyB2001icml,ZadroznyB2002kdd}, we can assume that
the class membership probabilities are given by $\bt_o  = [T_{j,o}] \in \Real^{K}$
where $T_{j,o}$ is defined to be 1 if the label of $\bx_o$ is $j$ and 0 otherwise. Then, MSE is calculated as
\be
	\mbox{MSE}(\bx_o) & = & \sum_{j=1}^K \Big(  T_{j,o} - \hat{P}(y_o=j|\bx_o) \Big)^2,
\ee
where $\hat{P}(y_o=j|\bx_o)$ is the class membership probability estimated by the method.
Note that MSE is also called Brier score \cite{BrierGW50mwr}, and the lower value the better performance.

For binary-decomposition methods, the base binary classifier was chosen according to the scale of dataset:
LibSVM for the datasets \{GCM, ALL\}, and Liblinear \cite{FanRE2008jmlr} for other datasets.
After learning of the binary classifiers, our method and WMAP were applied to learn the aggregation weights.
Due to computational complexity, WMAP was applied to the small datasets, \{GCM, ALL\}.
Otherwise, class membership probabilities for test dataset are
calculated in the WMAP framework with the fixed aggregation weights.
Each experiment was repeated 20 times by the random 10-fold cross-validation,
in which the original data are randomly split into 10 subsets with the equal size,
and then 1 subset is used for the validation data, and 9 subsets for the training data.

Table \ref{tab:acc_peformance_linear} summarizes the average accuracy for the different methods,
M-SVM, LMR, loss-based decoding, GBTM, WMAP, and our method.
The aggregation methods with APs encoding usually show the higher classification accuracy than two direct methods.
This result is consistent with the comparison study in \cite{HsuCW2002ieeetnn} which reported that
APs-based decomposition methods generally show higher predictive accuracy than
the direct methods for multiclass problems, such as M-SVM.
In addition, our aggregation method shows superior performance than other methods across most of cases.

Table \ref{tab:mse_peformance_linear} shows the average MSE,
which measures the quality of class membership probability estimate obtained by each method.
At first we can check out that our method significantly improves the quality of
class membership probability estimates of the loss-based decoding by tuning the aggregation weights.
In addition, LMR generally shows high performance for most data sets,
however, its performance does not exceed that of our method with APs encoding.
We finally conclude that our aggregation method outperforms other methods,
including the direct method, LMR, and two aggregation methods, GBTM and WMAP,
in terms of the quality of class membership probability estimates.
Note that, M-SVM is not considered in these cases due to its deterministic nature.

We also evaluated the performance of our method in terms of training time,
reported in Table \ref{tab:runtime_peformance}.
As mentioned before, WMAP is prohibitive even for medium-scale problems:
for GCM dataset, it averagely took 360.297 and 92.971 second
to learn the aggregation weights in AP and OVA encodings, respectively.
On the other hand, our method was terminated in seconds for most cases.
Furthermore, our method has an additional computational advantage over
the probabilistic decoding method based on (generalized) Bredely-Terry model, such as GBTM and WMAP,
which involve additional optimizations to estimate the class membership probabilities for test data.
In our method they are easily calculated by evaluating the softmax function
(\ref{eq:soft_max}) with the learnt aggregation weights.
For example, we compared the test time of 3 probabilistic decoding methods,
GBTM, WMAP and our method, on 6 UCI datasets.
As shown in Table \ref{tab:test_peformance_linear}, our method is remarkably faster than other methods.
As a results, we can confirm the superiority of our method
in terms of computational efficiency as well as classification performance.
Our method becomes more useful for large-scale multiclass problems
which involve evaluating class membership probabilities for the data points.

\begin{landscape}
\begin{table*}[htp]
\begin{center}
\caption{Comparison of classification performance for two direct methods (M-SVM and LMR),
and four aggregation methods (loss-based decoding, GBTM, WMAP, and our method),
in which results are the average accuracy and the number in parenthesis represents the standard deviation.
The numbers in bold face denote the best performance for each dataset.}
\label{tab:acc_peformance_linear}
\begin{tabular}{|c|c|c||c|c|c|c|c|c|}
\hline
       {Dataset}            &  M-SVM     &  LMR       &  Encoding  &  Loss-based
                            &    GBTM   & WMAP   & Our method    \\
\hline
   \multirow {3}*{GCM}      &             &             & APs       &  0.650(0.097)   &  0.632(0.092)  &  0.695(0.109) &  0.726(0.100) \\
                            &  0.708      &   0.684     & OVA       &  0.784(0.101)   &  0.784(0.101)  &  0.784(0.101) & \textbf{0.795(0.090)} \\
                            & (0.110)     &   (0.109)   & ECOC      &  0.758(0.096)   & 0.771(0.101)   &  0.721(0.123) & 0.766(0.100)   \\
\hline
  \multirow {3}*{ALL}       &             &             &  APs      & 0.978(0.033)    & 0.978(0.033)   &  0.978(0.033) & 0.972(0.035)   \\
                            & 0.972       &    0.976    & OVA       & 0.978(0.030)    & 0.978(0.030)   &  0.976(0.030) & 0.978(0.030)   \\
				            & (0.037)     &    (0.027)  & ECOC      & \textbf{0.980(0.028)}    & \textbf{0.980(0.028)}   &  0.978(0.027) & \textbf{0.980(0.028)}  \\
\hline
   \multirow {3}*{glass}    &             &             & APs       &  0.564(0.102)   & 0.576(0.094)   &  0.564(0.099) & 0.602(0.098)  \\
                            &  0.638      &  0.631      & OVA       &  0.600(0.098)   & 0.600(0.098)   &  0.598(0.102) & 0.610(0.108)  \\
                            &  (0.091)    &  (0.091)    & ECOC      &  0.629(0.109)   & 0.629(0.109)   &  0.626(0.111) & \textbf{0.640(0.088)} \\
\hline
  \multirow {3}*{segment}   &             &             & APs       & 0.950(0.017)    & 0.947(0.017)   &   0.948(0.017)   & \textbf{0.952(0.017)}  \\
                            &  0.950      &  0.907      & OVA       & 0.910(0.024)    & 0.910(0.024)   &   0.910(0.024)   & 0.917(0.022)\\
				            & (0.020)     &  (0.019)    & ECOC      & 0.905(0.022)    & 0.905(0.022)   &   0.905(0.022)   & 0.951(0.016)  \\
\hline
\multirow {3}*{satimage}    &             &             & APs       & \textbf{0.862(0.011)}    & 0.861(0.011)   &   0.861(0.011)   & \textbf{0.862(0.011)} \\
                            & 0.847       &  0.846      & OVA       & 0.832(0.014)    & 0.832(0.014)   &   0.832(0.014)   & 0.836(0.013) \\
                            & (0.011)     &  (0.015)    & ECOC      & 0.818(0.011)    & 0.818(0.011)   &   0.818(0.011)   & 0.856(0.013)  \\
\hline
\multirow {3}*{pendigits}  &              &             & APs       & 0.979(0.005)    & 0.977(0.006)   &  0.977(0.006)     & \textbf{0.979(0.005)}  \\
                           &  0.956       & 0.934       & OVA       & 0.931(0.007)    & 0.931(0.007)   &  0.931(0.007)     & 0.934(0.008)  \\
                           &  (0.007)     & (0.007)     & ECOC      & 0.918(0.025)    & 0.928(0.009)   &  0.928(0.009)     & 0.958(0.008)   \\
\hline
\multirow {3}*{isolet}     &              &             & APs       & 0.974(0.005)    & 0.973(0.005)   &  0.973(0.005)     & \textbf{0.978(0.004)}  \\
                           & 0.976    	  &  0.960      & OVA       & 0.972(0.005)    & 0.972(0.005)   &  0.972(0.005)     & 0.972(0.005)  \\
				           & (0.004)      &  (0.007)    & ECOC      & 0.959(0.008)    & 0.965(0.006)   &  0.965(0.006)     & 0.969(0.006) \\
\hline
\multirow {3}*{letter}     &              &             & APs       & 0.837(0.006)    & 0.830(0.007)   &   0.831(0.007)    & \textbf{0.844(0.007)}\\
                           & 0.784    	  &	0.748       & OVA       & 0.723(0.008)    & 0.723(0.008)   &   0.723(0.008)    & 0.723(0.009) \\
				           & (0.009)      & (0.010)     & ECOC      & 0.565(0.024)    & 0.619(0.016)   &   0.618(0.016)    & 0.635(0.027)  \\
  \hline
\end{tabular} 
\end{center}
\end{table*}
\end{landscape}

\begin{table*}[htp]
\begin{center}
\caption{Comparison of MSE for LMR and four aggregation methods (loss-based decoding, GBTM, WMAP, and our method).}
\label{tab:mse_peformance_linear}
{\small
\begin{tabular}{|c|c||c|c|c|c|c|c|}
\hline
       {Dataset}                  &  LMR      &  {Encoding}   &    Loss-based
                                  &    GBTM     & WMAP    & Our method \\
\hline
   \multirow {3}*{GCM}      & \multirow {3}*{0.476(0.084)}   & APs    &  0.923(0.001) & 0.569(0.068)  &  0.553(0.066)  & 0.417(0.148) \\
                                            &                & OVA    &  0.917(0.002) & 0.322(0.082)  &  0.347(0.079) & \textbf{0.287(0.125)} \\
                                            &                & ECOC   &  0.910(0.002) & 0.401(0.077)  &  0.423(0.084) & 0.340(0.136) \\
\hline
  \multirow {3}*{ALL}         & \multirow {3}*{0.068(0.041)} & APs    & 0.784(0.002)  & 0.067(0.048)  & 0.060(0.047)  & 0.039(0.050)  \\
                                			    &            & OVA    & 0.739(0.005)  & 0.042(0.041)  & 0.048(0.046)  & 0.038(0.049)  \\
				                            &                & ECOC   & 0.680(0.008)  & 0.043(0.044)  &  0.044(0.044) & \textbf{0.037(0.049)} \\
\hline
   \multirow {3}*{glass}    & \multirow {3}*{\textbf{0.516(0.116)}}   & APs   &  0.799(0.003)  & 0.552(0.092)& 0.550(0.089)   & 0.554(0.135) \\
                                           &                 & OVA    &  0.807(0.004) & 0.584(0.070)  &   0.585(0.070) & 0.568(0.113)  \\
                                           &                 & ECOC   &  0.791(0.006) & 0.568(0.068)  &  0.569(0.067) & 0.538(0.116) \\
\hline
  \multirow {3}*{segment} & \multirow {3}*{0.149(0.028)}     & APs    & 0.819(0.000)  & 0.107(0.018)  &  0.132(0.017) & \textbf{0.075(0.023)} \\
                                			    &            & OVA    & 0.798(0.002)  & 0.149(0.022)  &  0.153(0.022) & 0.112(0.023) \\
				                            &                & ECOC   & 0.758(0.003)  & 0.189(0.021)  &  0.190(0.021) & 0.077(0.020) \\
\hline
\multirow {3}*{satimage}& \multirow {3}*{0.206(0.013)}       & APs    & 0.783(0.000)  & 0.192(0.013)  & 0.196(0.012)  & \textbf{0.187(0.015)} \\
                                        &                    & OVA    &  0.764(0.001) & 0.242(0.012)  & 0.244(0.012)  & 0.224(0.015) \\
                                        &                    & ECOC   &  0.731(0.002) & 0.260(0.010)  & 0.261(0.010)  & 0.197(0.014) \\
\hline
\multirow {3}*{pendigits}& \multirow {3}*{0.110(0.009)}   & APs       & 0.880(0.000)  & 0.060(0.006)  &  0.141(0.006) & \textbf{0.033(0.007)} \\
                                        &                 & OVA       & 0.867(0.000)  & 0.121(0.009)  &  0.128(0.009) & 0.102(0.010) \\
                                        &                 & ECOC      & 0.860(0.002)  & 0.162(0.015)  &  0.178(0.014) & 0.065(0.011)  \\
\hline
\multirow {3}*{isolet}      & \multirow {3}*{0.085(0.007)} & APs      & 0.959(0.000)  & 0.228(0.006)  &  0.625(0.002) & \textbf{0.035(0.005)} \\
                                			    &          & OVA      & 0.956(0.000)  & 0.062(0.006)  &  0.165(0.007) & 0.042(0.005) \\
				                            &              & ECOC     & 0.945(0.001)  & 0.114(0.008)  &  0.171(0.008)  & 0.048(0.007)\\
\hline
\multirow {3}*{letter}      & \multirow {3}*{0.408(0.012)} & APs      & 0.959(0.000)  & 0.339(0.006)  &  0.663(0.002) & \textbf{0.232(0.007)}\\
                                			    &          & OVA      & 0.959(0.000)  & 0.495(0.006)  &  0.592(0.005) & 0.396(0.009)  \\
				                            &              & ECOC     & 0.955(0.000)  & 0.674(0.011)  &  0.697(0.010) & 0.505(0.026)  \\
  \hline
\end{tabular} }
\end{center}
\end{table*}

\begin{table*}[htp]
\begin{center}
\caption{Performance of our method in terms of training time,
in which results are the average iterations of the primal-dual interior point method required
to find the optimal solution and the average training time in second.}
\label{tab:runtime_peformance}
{\small
\begin{tabular}{|c|ccc||c|ccc|}
\hline
      { Dataset}         & { Encoding} &  {iter}  &  { Time}  &
      { Dataset}         & { Encoding} &  {iter}  &  { Time} \\
\hline
   \multirow {3}*{GCM}     &APs &  18.5  &  0.114(0.013)  & \multirow {3}*{satimage}&APs &  11.2  &  0.170(0.008)  \\
                           &OVA &  11.0  &  0.022(0.008)  &                         &OVA &  9.3   &  0.111(0.012)          \\
                           &ECOC&  12.6  &  0.053(0.007)  &                         &ECOC&  14.0  &  0.392(0.017)         \\
\hline
   \multirow {3}*{ALL}     &APs &  11.0  &  0.021(0.007)  & \multirow {3}*{pendigits} &APs & 13.8   &  1.350(0.044)  \\
                           &OVA &  9.0   &  0.015(0.007)  &                           &OVA & 9.0    &  0.313(0.013)        \\
                           &ECOC&  10.0  &  0.021(0.005)  &                           &ECOC& 14.9   &  1.544(0.046)       \\
\hline
   \multirow {3}*{glass}   &APs &  13.7  &  0.032(0.010)  & \multirow {3}*{isolet}    &APs & 23.5   &  41.879(1.490)  \\
                           &OVA &  10.9  &  0.017(0.009)  &                           &OVA & 13.5   &  1.412(0.059)        \\
                           &ECOC&  15.0  &  0.043(0.008)  &                           &ECOC& 14.3   &  3.903(0.132)       \\
\hline
   \multirow {3}*{segment} &APs &  12.8  &  0.132(0.122)  & \multirow {3}*{letter}    &APs & 23.5   &  106.299(2.249)  \\
                           &OVA &  10.0  &  0.052(0.002)  &                           &OVA & 15.0   &  3.916(0.104)    \\
                           &ECOC&  15.9  &  0.355(0.028)  &                           &ECOC& 16.6   &  11.201(0.397)    \\
\hline
\end{tabular} }
\end{center}
\end{table*}

\begin{table*}[htp]
\begin{center}
\caption{Comparison of test time for three probabilistic decoding methods
(GBTM, WMAP, and our method) on 6 UCI datasets,
in which results are the average test time in second and $N_0$ is the number of test points.  }
\label{tab:test_peformance_linear}
{\small
\begin{tabular}{|c||c|c|c|c|}
\hline
      Dataset     & Encoding  & GBTM   & WMAP  & Our method \\
\hline
                  & APs  &  0.178(0.026)&  0.174(0.006)&  0.000(0.000)     \\
     {glass}      & OVA  &  0.258(0.024)&  0.163(0.005)&  0.000(0.000)      \\
     ($N_0=21$)  & ECOC &  0.416(0.039)&  0.258(0.009)&  0.000(0.000)     \\
\hline
                  & APs  &  6.286(0.083)&  2.316(0.034)&  0.001(0.000)     \\
     {segment}    & OVA  &  5.234(0.170)&  2.264(0.052)&  0.001(0.000)    \\
     ($N_0=231$)  & ECOC &  6.700(0.242)&  5.000(0.032)&  0.004(0.000)     \\
\hline
                  & APs  &  15.331(0.134)&  6.398(0.114)&  0.001(0.000)      \\
     {satimage}   & OVA  &  8.652(0.270) &  6.027(0.055)&  0.002(0.000)     \\
     ($N_0=644$)  & ECOC &  11.776(0.582)&  9.247(0.032)&  0.004(0.000)     \\
\hline
                  & APs  &  43.769(0.152)&  14.356(0.115)&  0.007(0.000)     \\
     {pendigits}  & OVA  &  39.294(0.302)&  10.811(0.192)&  0.004(0.000)    \\
     ($N_0=1,099$)& ECOC &  45.466(0.957)&  28.057(0.873)&  0.008(0.001)     \\
\hline
                  & APs  &  88.248(0.554) &  26.034(0.156) &  0.032(0.002)     \\
     {isolet}     & OVA  &  68.645(0.210) &  9.614(0.305)  &  0.019(0.002)     \\
     ($N_0=780$)  & ECOC &  84.369(0.263) &  30.492(0.848) &  0.024(0.002)     \\
\hline
                  & APs  &  214.264(2.927)  & 94.640(1.006)    & 0.133(0.176)   \\
     {letter}     & OVA  &  167.560(6.766)  & 24.462(0.358)    & 0.051(0.005)    \\
     ($N_0=2,000$)& ECOC &  206.253(2.420)  & 60.249(0.727)    & 0.063(0.007)    \\
  \hline
\end{tabular} }
\end{center}
\end{table*}

We additionally evaluated the classification performance of
the aggregation methods with nonlinear binary classifiers.
We used the LibSVM with a rbf kernel function for a base binary classifier,
where the rbf-kernel width $\gamma$ and the regularization parameter $\lambda_B$ were
determined by maximizing the classification accuracy of randomly chosen validation set on
the 2-dimensional grid space ($\gamma, \lambda_B$), $\gamma\in[2^{-6},2^{-5},...,2^2,2^3]$ and
$\lambda_B \in[2^{-3},2^{-2},...,2^3,2^4]$.
Note that, the results on 2 cancer datasets (GCM and ALL) are not included in here
because the performance on these datasets were considerable worse than using the linear kernel.

The average classification accuracy and MSE are shown in Table \ref{tab:acc_peformance_kernel}
and \ref{tab:mse_peformance_kernel}, respectively.
As similar to the linear kernel case, our method showed
the stable performance for the wide range of the value of $\lambda$.
However, there was the slight degradation of performance of
our method on the datasets \{glass, satimage\} due to overfitting,
so we just increased the regularization parameter as $\lambda=10^{-1}$ for these datasets.
We can see that all probabilistic decoding methods yield the similar classification performance for most cases.
Although our method improves the results of the loss-based decoding in terms of MSE,
its performance is not significantly better than that of GBTM and WMAP.
The reason being is that optimal tuning of aggregation weights favors for good binary classifiers while de-emphasizing no good binary classifiers.
When suitably-chosen nonlinear kernels are used, all binary classifiers are already good, so no much performance gain is shown
even when optimal tuning of aggregation weights is made.

\begin{table*}[htp]
\begin{center}
\caption{Comparison of average classification accuracy for four aggregation methods
(loss-based decoding, GBTM, WMAP, and our method),
in the case where the nonlinear kernel (rbf function) is used for the SVM classifier.
The symbol ($\ast$) presented with the dataset name means
that we set the regularization parameter in our method, $\lambda$, to $10^{-1}$ for the given dataset.}
\label{tab:acc_peformance_kernel}
{\small
\begin{tabular}{|c|c||c|c|c|c|c|c|}
\hline
       {Dataset}            &  Encoding  &  Loss-based
                            &    GBTM  & WMAP  & Our method    \\
\hline
\multirow {3}*{glass ($\ast$)}    & APs    & 0.681(0.090)& 0.681(0.088)& \textbf{0.695(0.070)}& 0.683(0.094)  \\
                            & OVA          & 0.676(0.090)& 0.676(0.090)& 0.676(0.090)& 0.681(0.085)   \\
                            & ECOC         & 0.683(0.104)& 0.686(0.105)& 0.683(0.096)& \textbf{0.695(0.093)}  \\
\hline
\multirow {3}*{segment}            & APs   & 0.968(0.011)& 0.967(0.012)& 0.967(0.012)& 0.969(0.013)  \\
                            & OVA          & 0.971(0.012)& 0.971(0.012)& 0.971(0.012)& 0.971(0.012)  \\
                            & ECOC         & 0.975(0.009)& 0.975(0.009)& 0.975(0.009)& \textbf{0.977(0.010)}   \\
\hline
\multirow {3}*{satimage ($\ast$)}   & APs  & 0.924(0.009)& 0.924(0.009)& 0.924(0.009)& 0.924(0.010)    \\
                            & OVA         & 0.925(0.009)& 0.925(0.009)& 0.925(0.009)& 0.925(0.010)\\
                            & ECOC        & \textbf{0.927(0.010)}& \textbf{0.927(0.010)}& \textbf{0.927(0.010)}& \textbf{0.927(0.010)}   \\
\hline
\multirow {3}*{pendigits}  & APs       & 0.995(0.002)& 0.995(0.002)& 0.995(0.002)& 0.995(0.003) \\
                           & OVA       & \textbf{0.996(0.002)}& \textbf{0.996(0.002)}& \textbf{0.996(0.002)}& \textbf{0.996(0.002)}  \\
                           & ECOC      & 0.993(0.008)& \textbf{0.996(0.002)}& \textbf{0.996(0.002)}& \textbf{0.996(0.002)}   \\
\hline
\multirow {3}*{isolet}     & APs       & 0.971(0.004)& 0.970(0.005)& 0.970(0.005)& 0.977(0.004)   \\
                           & OVA       & 0.979(0.004)& 0.979(0.004)& 0.979(0.004)& 0.979(0.004) \\
				           & ECOC      & 0.980(0.004)& \textbf{0.981(0.003)}& \textbf{0.981(0.003)}& 0.979(0.004)  \\
\hline
\multirow {3}*{letter}     & APs       & 0.973(0.004)& 0.973(0.004)& 0.972(0.003)& 0.974(0.003)  \\
                           & OVA       & \textbf{0.978(0.004)}& \textbf{0.978(0.004)}& \textbf{0.978(0.004)}& 0.977(0.004)  \\
				           & ECOC      & 0.976(0.005)& \textbf{0.978(0.003)}& \textbf{0.978(0.003)}& \textbf{0.978(0.004)}   \\
  \hline
\end{tabular} }
\end{center}
\end{table*}

\begin{table*}[htp]
\begin{center}
\caption{Comparison of MSE for four aggregation methods (loss-based decoding, GBTM, WMAP, and our method),
in the case where the rbf-kernel function is used for the SVM classifier.}
\label{tab:mse_peformance_kernel}
{\small
\begin{tabular}{|c|c||c|c|c|c|c|c|}
\hline
       {Dataset}            &  Encoding  &  Loss-based
                            &    GBTM    & WMAP   & Our method    \\
\hline
\multirow {3}*{glass ($\ast$)}    & APs    &  0.795(0.002)& 0.441(0.081)& 0.447(0.064)&\textbf{0.440(0.082)} \\
                            & OVA          &  0.787(0.008)& 0.442(0.091)& 0.443(0.089)&0.442(0.102)   \\
                            & ECOC         &  0.762(0.010)& 0.426(0.083)& 0.447(0.079)&0.472(0.121)    \\
\hline
\multirow {3}*{segment}            & APs   & 0.818(0.000)& 0.059(0.015)& 0.075(0.014)&0.047(0.018)  \\
                            & OVA          & 0.783(0.001)& 0.043(0.015)& 0.044(0.014)&0.046(0.017)   \\
                            & ECOC         & 0.718(0.002)& \textbf{0.037(0.012)}& \textbf{0.037(0.012)}&0.039(0.017)   \\
\hline
\multirow {3}*{satimage ($\ast$)}  & APs  & 0.782(0.000)& 0.115(0.012)& 0.119(0.012)&0.124(0.011)  \\
                            & OVA         & 0.744(0.001)& 0.113(0.014)& 0.114(0.014)&0.115(0.014)  \\
                            & ECOC        &  0.692(0.002)& \textbf{0.111(0.013)}& \textbf{0.111(0.013)}&0.114(0.014)    \\
\hline
\multirow {3}*{pendigits}  & APs       & 0.881(0.000)& 0.018(0.003)& 0.089(0.003)&0.008(0.004)   \\
                           & OVA       & 0.860(0.000)& \textbf{0.006(0.003)}& 0.009(0.003)&\textbf{0.006(0.003)} \\
                           & ECOC      & 0.853(0.003)& 0.007(0.003)& 0.011(0.003)&\textbf{0.006(0.003)}    \\
\hline
\multirow {3}*{isolet}     & APs       &  0.959(0.000)& 0.115(0.007)& 0.563(0.002)&0.036(0.005) \\
                           & OVA       & 0.956(0.000)& 0.035(0.005)& 0.114(0.005)&0.034(0.005)  \\
				           & ECOC      & 0.944(0.001)& \textbf{0.032(0.004)}& 0.067(0.005)&0.035(0.006)   \\
\hline
\multirow {3}*{letter}     & APs       & 0.959(0.000)& 0.112(0.003)& 0.576(0.001)&0.040(0.005)  \\
                           & OVA       & 0.956(0.000)& 0.036(0.005)& 0.116(0.004)&\textbf{0.035(0.006)}   \\
				           & ECOC      & 0.943(0.001)& 0.037(0.004)& 0.073(0.005)&0.036(0.006)  \\
\hline
\end{tabular} }
\end{center}
\end{table*}

\section{Conclusions}
\label{sec:conclusions}

We have presented a method for optimally combining the results of binary classifiers
into a final answer to multiclass problems.
The softmax function was used to model the class membership probability,
taking a conic combination of discrepancies induced by binary classifiers and
returning a guess of class membership.
The corresponding log-likelihood was a convex function in the form of {\em log-sum-exp},
leading to a convex formulation for optimal binary classifier aggregation.
The primal-dual interior point method was adopted to solve the convex optimization problem.

Our method has several advantages over an existing optimal aggregation method,
WMAP \cite{YukinawaN2009ieeetcbb} which optimally combines binary class membership
probability estimates to form a joint probability estimates for all $K$ classes,
fitting the generalized Bradley-Terry model.
In WMAP, both aggregation weights and class membership probabilities are treated as
parameters to be estimated, so the computational complexity grows linearly with the number of
training examples. In contrast, our method has a few strong points:
(1) aggregation weights are the only parameters to be tuned (low complexity);
(2) the convex formulation yields the global solution;
(3) class membership probabilities for test data are easily evaluated without further optimizations.
In addition, our method is available for any types of discrepancy measures,
while the aggregation methods based on the (generalized) Bradley-Terry model
always require that the binary classifier yields the probability estimates.

The (primal-dual) interior point method still suffers from computational burden in large scale problems.
We may use  a stochastic approximation of interior point methods \cite{CarbonettoP2008nips}
to improve the scalability. It is our future work to adopt more efficient optimization to speed up the computation
and to improve the scalability, in our convex aggregation method.

\section{Appendix}
\subsection{Derivations of gradient and Hessian of the objective function (\ref{eq:convex_opt})}
\label{sec:app_grad_Hessian}

In this section, we include the gradient and Hessian of the objective function (\ref{eq:convex_opt}),
which can be easily calculated based on the derivations of gradient and Hessian of the log-sum-exp function,
described in Appendix in \cite{BoydS2004book}.

We first compute $\bvarphi_i^{j,y_i}$ for $j=1,...,K$ and $i=1,...,N$ by (\ref{eq:varphi}).
Then we define $\bPsi_i \in \Real^{K\times M}$ and
$\bu_i \in \Real^{K\times 1}$ for the $i$th data point:
\be
    \bPsi_i  & = &
        \begin{bmatrix}
            (\bvarphi_i^{1,y_i})^\top \\
            \vdots \\
            (\bvarphi_i^{K,y_i})^\top
        \end{bmatrix}. \nonumber \\
    \bu_i & = & \left[\exp\{\bw^\top\bvarphi_i^{1,y_i}\},...,
        \exp\{\bw^\top\bvarphi_i^{M,y_i}\} \right]^\top.
\ee
The objective function is given by
\be
    f_0(\bw) = \frac{1}{N}\sum_{i=1}^{N} \log \left( \sum_{k=1}^{K}
               [\bu_i]_k \right) + \frac{\lambda}{2}\bw^\top\bw.
\ee
The gradient is computed as
\be
    \nabla f_0(\bw) = \sum_{i=1}^N \left(\frac{1}{\mathbf{1}^\top\bu_i}
        \bPsi_i^\top\bu_i\right) + \lambda \bw. \nonumber
\ee
The Hessian is obtained by
\be
    \lefteqn{\nabla^2 f_0(\bw)}  \nonumber\\
        & = & \sum_{i=1}^N \bPsi_i^\top\left(\frac{1}{\mathbf{1}^\top\bu_i}\diag(\bu_i)
        -\frac{1}{(\mathbf{1}^\top\bu_i)^2}\bu_i(\bu_i)^\top\right)\bPsi_i + \mbox{diag}(\blambda), \nonumber
\ee
where $\blambda = [\lambda,...,\lambda]^\top \in \Real^{M}$.

\subsection{Proof of Proposition \ref{prop:p1}}
\label{sec:app_proof1}
{\em Proof.}
Let us define $\l_{\tau}$ as the modified log-sum-exp function with $\tau$, such that
\be
\lefteqn{ \l_{\tau}\left(\bvarphi_i^{1,y_i},\bvarphi_i^{2,y_i},...,\bvarphi_i^{K,y_i},\bw\right) } \nonumber \\
&= & \frac{1}{\tau}
        \log \left( \sum_{k=1}^K \exp \Big\{ \tau\Big((1-\delta(y_i,k)) + \bw^\top \bvarphi_i^{k,y_i} \Big) \Big\} \right).
        \nonumber
\ee
We first show that the sequence of functions
$\left\{ \l_{\tau}\left(\bvarphi_i^{1,y_i},\bvarphi_i^{2,y_i},...,\bvarphi_i^{K,y_i},\bw\right) \right\}$
for $\tau=1,2,\ldots,$
uniformly converges to the multiclass hinge loss function
$h \left(\bvarphi_i^{1,y_i}, \bvarphi_i^{2,y_i}, \ldots, \bvarphi_i^{K,y_i}, \bw \right)$.
Then, we can easily prove that the sequence of functions $\{f_\tau(\bw)\}$ also uniformly converges to $f_{LM}(\bw)$.
This proposition is an extension of the results in \cite{ZhangJ2003icml} which provide a connection between
the hinge loss function and logistic loss function (that is a special case of the log-sum-exp function)
in the case of binary problems.

For all $\bxi\in \Real^{K}$, we are given the following inequalities for the log-sum-exp function \cite{BoydS2004book}:
\be
    \label{eq:inequality_logsumexp1}
    \max\{ \xi_1,\xi_2,...,\xi_K \} & \leq & \log \left(\sum_{k=1}^K \exp\{\xi_k\} \right) \nonumber \\
    & \leq & \max\{ \xi_1,\xi_2,...,\xi_K \} + \log K.
\ee
Substituting $\xi_k = \tau\Big((1-\delta(y_i,k)) + \bw^\top \bvarphi_i^{k,y_i} \Big)$ into (\ref{eq:inequality_logsumexp1}),
one can easily see that
\be
    \label{eq:inequality_logsumexp2}
\lefteqn{h \left(\bvarphi_i^{1,y_i}, \bvarphi_i^{2,y_i}, \ldots, \bvarphi_i^{K,y_i}, \bw \right)}
\nonumber \\
& \leq &
         \l_{\tau}\left(\bvarphi_i^{1,y_i},\bvarphi_i^{2,y_i},...,\bvarphi_i^{K,y_i},\bw\right)  \nonumber\\
    & \leq & h \left(\bvarphi_i^{1,y_i}, \bvarphi_i^{2,y_i}, \ldots, \bvarphi_i^{K,y_i}, \bw \right) + \frac{\log K}{\tau}.
\ee
It follows from this inequality that we have
\be
   \frac{\log K}{\tau} & = & \max_{ \bw, \bvarphi_i^{1,y_i} ,.., \bvarphi_i^{K,y_i}}
    \left\{ \l_{\tau}\left(\bvarphi_i^{1,y_i},\bvarphi_i^{2,y_i},...,\bvarphi_i^{K,y_i},\bw\right) \right.\nonumber \\
  &&   \left.   - h \left(\bvarphi_i^{1,y_i}, \bvarphi_i^{2,y_i}, \ldots, \bvarphi_i^{K,y_i}, \bw \right) \right\}.
\ee
Thus, for any given $\epsilon > 0$, we can choose sufficiently large $\tau$ such that
\bee
 \left|\l_{\tau}  \left(\bvarphi_i^{1,y_i},\bvarphi_i^{2,y_i},...,\bvarphi_i^{K,y_i},\bw\right)
  - h \left(\bvarphi_i^{1,y_i}, \bvarphi_i^{2,y_i}, \ldots, \bvarphi_i^{K,y_i}, \bw \right) \right|
         < \epsilon.
\eee
Summing over all training data points, we obtain the inequality
${\log K}/{\tau} \geq \max_{\bw}\left\{ f_\tau(\bw) - f_{LM}(\bw) \right\}$,
which implies the uniform convergence of $f_\tau(\bw)$ to $f_{LM}(\bw)$.

\subsection{Proof of Proposition \ref{prop:p2}}
\label{sec:app_proof2}
{\em Proof.}
We first give the detailed explanations for some notations used in the proof.
We can consider the margin of each example, $\nu_{w}(\bx_i,y_i)$, as a decision function for multiclass problems.
For example, the data point $\bx_i$ is misclassified if and only if $\nu_w(\bx_i,y_i) \leq 0$.
Thus the empirical misclassification error can be defined by
\be
   \frac{1}{N}\sum_{i=1}^N  \bone\big( \nu_w(\bx_i,y_i) \leq 0 \big),
\ee
where $\bone(\pi)$ is the {\em 0-1 loss function} which equals 1 if the predicate $\pi$ is true, otherwise 0.
In addition, due to the definition of the margin, we have
\be
    \label{eq:margin}
    \nu_w(\bx_i,y_i) & = & \min_{k \neq y_{i}}\rho(\bc_k,\bq_i,\bw) - \rho(\bc_{y_i},\bq_i,\bw) \nonumber\\
        & = & \bw^\top \bvarphi_i^{\bar{k}_i,y_i}
\ee
where $\bar{k}_i = \argmin_{k \neq y_{i}}\rho(\bc_k,\bq_i,\bw)$.
Note that, since $\bvarphi^{\overline{k},y}_{x}$ can be considered as feature mapping, as in kernel methods,
the hypothesis set considered here is defined as a bounded linear function class,
i.e., $\calF = \{f: \calX \times \calY \rightarrow \Real ~|~ f(\bx,y)
= \bw^\top\bvarphi_i^{\bar{k}_i,y_i} \mbox{ for some } \bw \in \Real_+^M, \| \bw\|_2 \leq B\}$.
In this work, we aim at finding an aggregation weight vector among the function class $\calF$,
which minimizes the expected misclassification error (generalization error),
$\mathbb{E}\left[ \bone\Big( \nu_w(\bx,y) \leq 0\Big) \right]$ $\big(= P(y \neq \hat{y}) \big)$.
However, a direct optimization involving the 0-1 loss is not an easy task because of its discrete nature.
We instead consider a ramp loss $\phi: \Real \rightarrow [0, \, 1]$,
which is a continuous upper bound on {\em 0-1 loss function}:
\be
    \label{eq:smooth_hinge}
    \phi(z) =
        \left\{ \begin{array}{rl}
          0        & \mbox{ if $z\geq 1$} \\
          1 - z    & \mbox{ if $0<z<1$} \\
          1        & \mbox{ if $z\leq 0$} \\
        \end{array} \right..
\ee
Note that, $\phi$ is a clipped version of hinge loss \cite{CollobertR2006jmlr}.
Finally, we define the empirical Rademacher complexity \cite{BartlettPL2004jmlr}
of a class of functions we are interested in.
Let $\calG$ be a class of functions mapping from $\calX \times \calY$ to $\Real$
and given samples $\{(\bx_i,y_i)\}_{i=1}^N$,
the empirical Rademacher complexity of the class $\calG$ is given by \cite{BartlettPL2004jmlr}
\be
    \widehat{\mathfrak{R}}_N(\mathcal{\calG}) = \mathbb{E}_{\sigma}\Bigg[\sup_{h \in \calG}
        \Big(\frac{1}{N}\sum_{i=1}^N\sigma_i h(\bx_i,y_i) \Big)\Bigg],
\ee
where $\sigma_i \in \{-1, 1\}$ are independent uniform random variables.
Note that, the empirical Rademacher complexity is based on
the training examples and thus is practically computable.
In addition, it can be viewed as the correlation between a random binary noise
and functions in the function class $\calG$, in the supermum sense.
In our case, the empirical Rademacher can be calculated based on Lemma 22 in \cite{BartlettPL2004jmlr}.
Defining a new index $k_i' = \argmin_{k\neq y_i} \| \bvarphi_i^{k,y_i} \|_2$,
the empirical Rademacher complexity of the class $\calF$ is given by
\be
    \label{eq:emp_Rademacher_com}
    \widehat{\mathfrak{R}}_N(\mathcal{F})
        & = & \mathbb{E}_{\sigma}\Bigg[\sup_{\bw : \|\bw\|_2 \leq B \mbox{ and } \bw \in \Real^M_+}
            \bw^\top\Big( \frac{1}{N}\sum_{i=1}^N\sigma_i \bvarphi_i^{\bar{k}_i,y_i} \Big)\Bigg] \nonumber\\
        & \leq & \frac{B}{N}\mathbb{E}_{\sigma}\left[\sqrt{\sum_{i=1}^N\sum_{j=1}^N\sigma_i\sigma_j
            \big(\bvarphi_i^{k_i',y_i} \big)^\top\bvarphi_j^{k_i',y_j}}\right] \nonumber\\
        & \leq & \frac{B}{N}\sqrt{\mathbb{E}_{\sigma}\left[\sum_{i=1}^N\sum_{j=1}^N\sigma_i\sigma_j
            \big(\bvarphi_i^{k_i',y_i} \big)^\top\bvarphi_j^{k_i',y_j}\right]}  \nonumber\\
        & = & \frac{B}{N} \Bigg(\sum_{i=1}^N \min_{k\neq y_i}\|\bvarphi_i^{k,y_i}\|_2^2\Bigg)^{{1}/{2}},
\ee
where Cauchy-Schwarz and Jensen's inequalities are used to
arrive at the second and the third inequalities respectively.

Applying Theorem 7 and 8 in \cite{BartlettPL2004jmlr}, yields the following generalization bound.
With the empirical Rademacher complexity in (\ref{eq:emp_Rademacher_com}),
for any $\epsilon>0$, with probability greater than $1 - \epsilon$ over samples of length $N$,
every aggregation weight vector $\bw\in \calF$ satisfies
\be
    \label{eq:bound}
     \mathbb{E}\left[ \bone\Big( \nu_w(\bx,y) \leq 0\Big)  \right] & \leq &
        \frac{1}{N} \sum_{i=1}^N \phi \Big( \nu_w(\bx_i,y_i) \Big)
         + 2\,\widehat{\mathfrak{R}}_N(\mathcal{\calF}) + \sqrt{\frac{9\ln(2/\epsilon)}{2N}}.
\ee
Using $P(y \neq \hat{y})=\mathbb{E}\left[ \bone\Big( \nu_w(\bx,y) \leq 0\Big) \right]$,
we directly obtain Proposition 2.

\bibliographystyle{unsrtnat}
{\bibliography{sjc}}

\end{document}